\documentclass[10pt,twocolumn,letterpaper]{article}

\usepackage{pifont}
\usepackage{algorithm}
\usepackage{adjustbox}
\usepackage{array}
\usepackage{multirow}
\usepackage{xspace}
\usepackage{pifont}
\usepackage[noend]{algpseudocode}
\usepackage{makecell}
\usepackage{caption}
\usepackage{booktabs}
\usepackage{bm}

\usepackage{relsize}

\usepackage{amsmath, amsthm, amssymb}
\theoremstyle{plain}
\newtheorem{theorem}{Theorem}[section]
\theoremstyle{observation}

\usepackage{cvpr}              

\definecolor{cvprblue}{rgb}{0.21,0.49,0.74}
\usepackage[pagebackref,breaklinks,colorlinks,allcolors=cvprblue]{hyperref}
\usepackage{graphicx}
\def\paperID{9094} 
\def\confName{CVPR}
\def\confYear{2025}

\title{Invisible Backdoor Attack against Self-supervised Learning}

\author{
  Hanrong Zhang$^1$\thanks{Equal Contribution. Email: \href{zhanghr0709@gmail.com}{zhanghr0709@gmail.com};\\  \href{zhenting.wang@rutgers.edu}{zhenting.wang@rutgers.edu}} 
  \quad
  Zhenting Wang$^2$\footnotemark[1] 
\quad
  Boheng Li$^3$
  \quad
  Fulin Lin$^1$
  \quad
  Tingxu Han$^4$ \\
  \quad
  Mingyu Jin$^2$
  \quad
  Chenlu Zhan$^1$
  \quad
  Mengnan Du$^5$
  \quad
  Hongwei Wang$^1$\thanks{Corresponding author. Email: \href{mailto:hongweiwang@intl.zju.edu.cn}{hongweiwang@intl.zju.edu.cn}} 
  \quad
  Shiqing Ma$^6$ \\
  $^1$ Zhejiang University \quad $^2$ Rutgers University \quad $^3$ Nanyang Technological University \\ $^4$ Nanjing University \quad $^5$ New Jersey Institute of Technology \quad $^6$ University of Massachusetts Amherst
}

\begin{document}
\maketitle


\begin{abstract}
Self-supervised learning (SSL) models are vulnerable to backdoor attacks. Existing backdoor attacks that are effective in SSL often involve noticeable triggers, like colored patches or visible noise, which are vulnerable to human inspection. This paper proposes an imperceptible and effective backdoor attack against self-supervised models. We first find that existing imperceptible triggers designed for supervised learning are less effective in compromising self-supervised models. We then identify this ineffectiveness is attributed to the overlap in distributions between the backdoor and augmented samples used in SSL. Building on this insight, we design an attack using optimized triggers disentangled with the augmented transformation in the SSL, while remaining imperceptible to human vision. Experiments on five datasets and six SSL algorithms demonstrate our attack is highly effective and stealthy. It also has strong resistance to existing backdoor defenses. Our code can be found at \url{https://github.com/Zhang-Henry/INACTIVE}.
\end{abstract}

\section{Introduction}
\label{sec:intro}

In recent years, Self-Supervised Learning (SSL) has become a powerful approach in deep learning, enabling the learning of rich representations from vast unlabeled data, thus avoiding manual labeling. SSL aims to develop an image encoder that produces similar embeddings for similar images by applying various \emph{augmentations} to the same image. This pre-trained encoder can be used for different downstream tasks by training compact downstream classifiers with relatively few parameters.

Although SSL has been extensively used in the development of foundational models~\cite{chen2020simple,he2020momentum,Caron_2021_ICCV}, it is at risk of backdoor attacks~\cite{Jia_Liu_Gong_2022,Saha_Tejankar_Koohpayegani_Pirsiavash_2022,Li_2023_ICCV,tao2023distribution}, where the attacker embeds hidden malicious behavior within the encoder.
The backdoor can be inherited to the downstream task. The downstream classifier predicts a specific target label if the input contains a pre-defined backdoor trigger.
Existing backdoor attacks on SSL such as BadEncoder~\cite{Jia_Liu_Gong_2022} achieve high attack success rates (ASR). 
However, \emph{a common drawback of these effective attacks is that their trigger patterns are obvious, making them susceptible to human inspection.} 
Moreover, while data-poisoning-based attacks CTRL~\cite{Li_2023_ICCV} and BLTO~\cite{sun2024backdoor} are relatively stealthy, their ASRs are suboptimal, For example,  on CIFAR10 CTRL only has 61.90\% ASR under BYOL framework and BLTO only has 84.63\% ASR under SimSiam framework. Furthermore, they also rely on the downstream dataset matching the pre-training dataset distribution, limiting effectiveness across diverse datasets.
In this paper, we aim to propose a backdoor attack in SSL that is both effective and stealthy to human vision without this distribution dependency.

\begin{figure}[t]
  \centering
   \includegraphics[width=\linewidth]{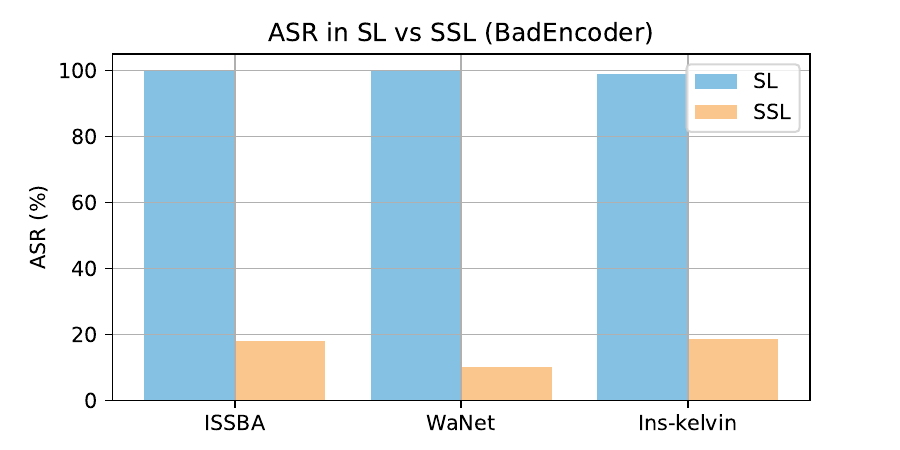}

   \caption{Existing imperceptible backdoor triggers, which yield high ASR in supervised learning (SL), do not perform as effectively in SSL. The attack framework for SL and SSL are standard backdoor poisoning~\cite{gu2017badnets} and BadEncoder~\cite{Jia_Liu_Gong_2022}, respectively.}
   \label{fig:asr}
   \vspace{-15pt}
\end{figure}

There are various invisible triggers designed for the backdoor attacks on supervised classifiers, such as WaNet~\cite{nguyen2021wanet}, ISSBA~\cite{li2021invisible}, and filter attack~\cite{liu2019abs}. A straightforward way to achieve imperceptible backdoor attacks in SSL is by directly applying these invisible triggers. However, these existing invisible triggers designed for supervised learning do not perform as well in attacking self-supervised models (see \autoref{fig:asr}). We then find that 
this lack of effectiveness is due to the overlapping distributions between the backdoor samples and the augmented samples utilized in SSL. Namely, self-supervised models cannot effectively distinguish the distribution of the backdoor samples and the augmented samples, due to the similarity between the transformation altered by the backdoor trigger and intrinsic image augmentations in SSL, such as RandomGrayscale and ColorJitter (see \autoref{fig:observation}).

\begin{figure*}[t]
  \centering
   \includegraphics[width=\linewidth]{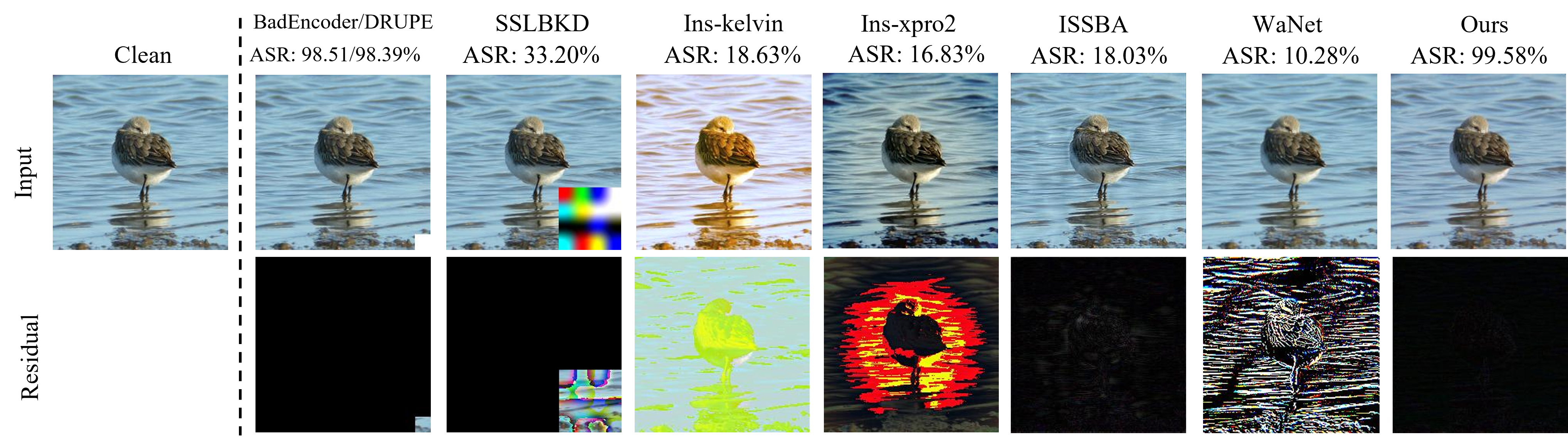}

   \caption{Comparison of clean, backdoored samples created by Patch trigger used by BadEncoder~\cite{Jia_Liu_Gong_2022} and DRUPE~\cite{tao2023distribution}, Instagram filter trigger~\cite{pilgram}, ISSBA trigger~\cite{li2021invisible}, WaNet trigger~\cite{nguyen2021wanet} and ours. Except for DRUPE, the ASRs are tested under the threat model of BadEncoder. 
    Residuals are the difference between clean and backdoored images. \emph{Our method achieves the highest ASR while maintaining trigger stealthiness, while other methods either have a much lower ASR or use more easily detectable triggers.}}
   \label{fig:compare}
   \vspace{-15pt}
\end{figure*}

Based on the above observations, we developed a backdoor attack that disentangles its optimized trigger transformation and the augmented transformation in SSL. In detail, it involves increasing the distributional distance between backdoor samples and the augmented samples in the SSL process. We also keep the trigger stealthy by adding the constraints on both pixel-space and feature-space distance to the original samples, using metrics like SSIM, PSNR, perceptual loss, and Wasserstein distance. We then implemented our prototype 
INACTIVE (\textbf{IN}visible b\textbf{AC}kdoor a\textbf{T}tack \textbf{I}n self-super\textbf{V}ised l\textbf{E}arning) and 
tested it on five datasets (i.e., CIFAR10, STL10, GTSRB, SVHN, ImageNet), and across six classic SSL frameworks  (i.e., SimCLR~\cite{chen2020simple}, MoCo~\cite{he2020momentum}, BYOL~\cite{grill2020bootstrap}, SimSiam~\cite{chen2021exploring}, SwAV~\cite{caron2020unsupervised} and CLIP~\cite{radford2021learning} (See \autoref{tab:comparison} and \autoref{tab:SSLs}) with their various augmentation transforms (See \autoref{tab:parameters}). The results demonstrate that our method is highly effective and stealthy. In detail, it achieves an average of 99.09\% ASR (See \autoref{tab:comparison}), 0.9763 SSIM, 41.07 PSNR, 0.0046 LIPIS, 0.9751 FSIM, and 13.281 FID (See \autoref{tab:stealthiness_detail}). 
As shown in \autoref{fig:compare}, we compare several methods' backdoor residuals and ASRs. Our method exhibits the highest ASR while maintaining the highest stealthiness.
It also effectively bypasses existing backdoor defenses such as DECREE~\cite{Feng_2023_CVPR}, Beatrix~\cite{ma2022beatrix}, ASSET~\cite{ASSET}, STRIP~\cite{gao2019strip}, Grad-CAM~\cite{Selvaraju_2017_ICCV}, Neural Cleanse~\cite{Wang_Yao_Shan_Li_Viswanath_Zheng_Zhao_2019}, and various noise, i.e., JPEG compression, Poisson noise, and Salt\&Pepper noise. 

Our contributions are summarized as follows: 
\ding{172} We observed that existing imperceptible triggers designed for supervised classifiers have limited effectiveness in SSL. \ding{173} We find that the reason behind such ineffectiveness is the coupling feature-space distributions for the backdoor samples and augmented samples in the SSL models. \ding{174} Based on our findings, we propose an imperceptible and effective backdoor attack in SSL by disentangling the distribution of backdoor samples and augmented samples in SSL, while constraining the stealthiness of the triggers during the optimization process.
\ding{175} Extensive experiments on five datasets and six SSL algorithms with different augmentation ways demonstrate our attack is effective and stealthy, and can also be resilient to current SOTA backdoor defense methods.

\section{Related Work}
\subsection{Self-Supervised Learning}

The goal of SSL is to leverage a large amount of unlabeled data in the pre-training dataset to pre-train an image encoder, which can then be used to create classifiers for various downstream tasks with a smaller set of labeled data~\cite{tao2023distribution}. SSL pipelines for contrastive learning typically include the following approaches~\cite{Jaiswal_ramesh_babu_Zadeh_Banerjee_Makedon_2020, gui2023survey}:
\ding{172} Negative Examples: Promotes proximity among positive examples while maximizing the distance between negative examples in the latent space, as seen in SimCLRs~\cite{chen2020simple} and MoCo~\cite{he2020momentum}.
\ding{173} Self-distillation: Utilizes two identical Siamese networks with different weights to increase the similarity between differently augmented versions of the same image, such as in BYOL~\cite{grill2020bootstrap} and SimSiam~\cite{chen2021exploring}.
\ding{174} Clustering: Implements a clustering mechanism with swapped prediction of representations from both encoders, as in SwAV~\cite{caron2020unsupervised}.
\emph{Our method is shown to be highly effective and stealthy under the SSL algorithms in \autoref{tab:comparison} and \autoref{tab:SSLs}.}

\subsection{Backdoor Attacks}
Backdoor attacks were initially proposed for supervised learning (SL) to modify a model’s behavior on specific inputs or classes while keeping its general performance intact~\cite{li2024nearest,wang2022bppattack,mei-etal-2023-notable,tao2022backdoor,zhang2025agent,qi2022towards,zeng2023narcissus,gao2024backdoor,cai2024toward}. Early backdoor attacks commonly utilized visible triggers like distinctive patches that are easily detectable through visual inspection~\cite{gu2017badnets,shafahi2018poison,tang2021demon}. To enhance stealth, subsequent research introduced invisible triggers, which are subtle and blend into the background, helping these attacks evade both human inspection and certain automated defenses~\cite{li2021invisible,nguyen2021wanet}.

Since many of these attacks rely on labeled data, recent studies have explored alternative backdoor implantation techniques in SSL models~\cite{Saha_Tejankar_Koohpayegani_Pirsiavash_2022,carlini2022poisoning,liu2022poisonedencoder, Jia_Liu_Gong_2022,sun2024backdoor}. However, these typically use a visible backdoor trigger, such as a patch, making them prone to human detection and model simulation.
The advantage of invisible triggers in SSL is clear: they improve attack stealth, bypassing some conventional defenses that focus on detecting visible anomalies~\cite{Feng_2023_CVPR,tejankar2023defending}. 
However, as we will demonstrate, directly applying existing invisible triggers designed for SL to SSL tasks results in limited attack effectiveness.
Moreover, backdoor attacks in SSL are generally divided into two types: training-time backdoor injection attacks like BadEncoder~\cite{Jia_Liu_Gong_2022}, which require control of the training in the backdoor injection process, and data-poisoning-based attacks like CTRL~\cite{Li_2023_ICCV} and BLTO~\cite{sun2024backdoor}, which rely on poisoned data without needing model specifics.
BadEncoder modifies a pre-trained encoder to embed triggers that cause targeted misclassifications in downstream tasks. By aligning the features of triggered images with an attacker-chosen class, downstream classifiers misclassify triggered inputs while maintaining accuracy on clean data.
Although CTRL and BLTO are relatively stealthy, they achieve lower ASRs than our method and depend on the downstream dataset matching the pre-training distribution, limiting their versatility across diverse datasets.
In this paper, we focus on training-time backdoor injection attack due to it has higher attack effectiveness and transferability.



\subsection{Backdoor Defenses}

Various defenses have been developed against backdoor attacks~\cite{li2024purifying,xie2023badexpert,hou2024ibd,liu2022complex,wang2022unicorn,qi2023towards,qi2022revisiting,pan2023asset,zeng2023meta,chen2025refine,xu2024towards}, primarily targeting supervised classifiers. These defenses either prevent attacks during training~\cite{wang2022training,huang2022backdoor,hong2020effectiveness,tran2018spectral} or detect and mitigate backdoors in compromised models offline~\cite{tao2022better,zhu2023selective,tao2022model,wang2022rethinking,xu2021detecting,shen2021backdoor,liu2019abs,li2021neural}. Some methods also detect backdoor-triggered inputs during inference~\cite{gao2019strip}. 
Defense methods like DECREE~\cite{Feng_2023_CVPR}, Beatrix~\cite{ma2022beatrix}, and ASSET~\cite{ASSET} are designed for SSL, primarily relying on the visible characteristics of triggers to detect backdoors. In contrast, our method uses invisible triggers, effectively bypassing these defenses by breaking their reliance on visual anomalies and making detection more challenging.

\section{Observations and Analysis}

\label{sec:obs}



\begin{figure}[t]
  \centering
   \includegraphics[width=0.55\linewidth]{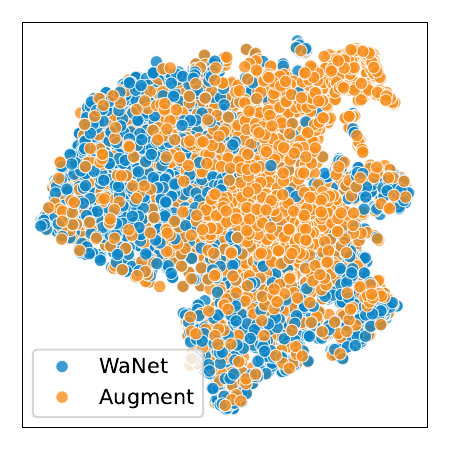}

\caption{t-SNE visualization of the feature space in the inherent augmentation and backdoor trigger space. The SimCLR~\cite{chen2020simple} pre-trained model struggled to differentiate between backdoor samples injected with the WaNet trigger~\cite{nguyen2021wanet} and the augmented samples within the SimCLR contrastive learning framework.}

   \label{fig:observation}
   \vspace{-15pt}

\end{figure}

\noindent
\textbf{Invisible Trigger Designed for SL Fails in SSL.}
We first assess the effectiveness of existing invisible triggers (WaNet, ISSBA, and filter attack) designed for supervised classifiers. \autoref{fig:asr} displays their ASRs on both supervised classifiers and self-supervised models. For supervised classifiers, the standard backdoor poisoning method~\cite{gu2017badnets} is used. For self-supervised models, we apply the BadEncoder method, replacing the patch trigger with these invisible triggers, using ResNet18. We find that these triggers, which achieve high ASR in supervised learning, are less effective in SSL. We then investigate the underlying reason for these results and focus on the following research question: \emph{Why does the effective invisible backdoor trigger designed for supervised learning fail on self-supervised learning?}

\noindent
\textbf{Cause of the Failure: Entanglement of the Inherent Augmentations and Backdoor Trigger.}
We find that the entanglement of the inherent augmentations in contrastive learning can cause the failure of the backdoor injection with such triggers.
We provide our analysis in this section.
One of the core training losses of contrastive learning can be formulated as maximizing the feature space similarity between the augmented samples modified from the same training samples:

\begin{equation}
\arg\max_{\theta_\mathcal{F}} s(\mathcal{F}_\theta(\mathcal{A}_1 (\bm x)),\mathcal{F}_\theta(\mathcal{A}_2 (\bm x)))
\label{eq:ssl_loss}
\end{equation}

\noindent
where \( s(\cdot,\cdot) \) denotes the similarity measurement,  \(\mathcal{F}_\theta\) is the encoder in training (\(\theta_\mathcal{F}\) is its parameters), \(\bm x\) is the training sample, \(\mathcal{A}_1\) and \(\mathcal{A}_2\) are different augmentations sampled from the \emph{predefined augmentation space} \(\mathcal{S}_\mathcal{A}\). 
Different from the predefined augmentation space, we also define the \emph{learned augmentation space} for trained encoders \(\mathcal{S}_\mathcal{A}^{\prime}\) as the space including a set of transformations where any pair within it can achieve high pairwise similarity on augmented versions of the same sample when processed by the trained encoder.
We also define \emph{perfectly-trained encoder} is the encoder that achieves maximal similarity described in \autoref{eq:ssl_loss} for all samples and all possible transformations used, \ie, \(s(\mathcal{F}_\theta(\mathcal{A}_1 (\bm x)),\mathcal{F}_\theta(\mathcal{A}_2 (\bm x))) = 1, \forall \bm x \in \mathcal{X}, \forall \mathcal{A}_1, \mathcal{A}_2 \in \mathcal{S}_\mathcal{A}\), where \(\mathcal{X}\) is the input space.
Based on this, we have the following theorem:

\begin{theorem}\label{th:perfect_trojan1}
    Given a perfectly-trained encoder \(\mathcal{F}_\theta\)  based on the augmentations sampled from predefined augmentation space \(\mathcal{S}_\mathcal{A}\), it is impossible to inject a backdoor with trigger function \(\mathcal{I} \in \mathcal{S}_\mathcal{A}\).
\end{theorem} 

\noindent
The proof of this theorem can be found in the \autoref{sec:proof}. In practice, the boundary of the learned augmentation space for trained encoders \(\mathcal{S}_\mathcal{A}^{\prime}\) is often imprecise, and it potentially reflects a relaxation of the predefined augmentation space. Consequently, using trigger functions that are not precisely within the predefined augmentation space 
\(\mathcal{S}_\mathcal{A}\) but are instead distributionally close to it can also make achieving high attack success rates hard.

\noindent
\textbf{Empirical Evidence.}
We also conduct experiments to confirm the invisible triggers designed for supervised learning are actually entangled with the inherent augmentations in self-supervised learning.
Specifically, we use a ResNet18 pre-trained with SimCLR for a binary classification task to differentiate between samples poisoned by WaNet and those augmented by SimCLR. We ensure consistent feature representations for clean samples between the backdoored and clean models using utility loss from Jia et al~\cite{jin2023backdoor}.
Results indicate that the models struggle to differentiate between the two categories. A t-SNE visualization of their features, as presented in \autoref{fig:observation}, indicates a significant overlap and entanglement. From this, we infer that the diminished effectiveness of supervised backdoor attack methods in SSL scenarios is attributed to the distributional similarity between the features of contrastive-learning-augmented samples and backdoor samples. 
The reason for the entanglement phenomenon on the invisible backdoor trigger is that the learned augmentation space reflects a relaxation of the predefined augmentation space, and such relaxation covers most of the invisible minor transformations. Thus, we aim to search for invisible transformations that can escape the inherent augmentation space.

\section{Attack Design}

\subsection{Threat Model}
\label{Threat_Model}

We follow the well-defined training-based backdoor injection threat model introduced in BadEncoder~\cite{Jia_Liu_Gong_2022}.

\noindent
\textbf{Objectives of the Attacker.}\
The objective of an attacker is to implant backdoors into a pre-trained image encoder by SSL. We define a backdoored image encoder model as $\mathcal{F}_\theta$ and the backdoor injector as $\mathcal{I}_\phi$.
In this way, a downstream classifier trained based on $\mathcal{F}_\theta$, which we define as $\mathcal{C}_\epsilon$, could produce a specific prediction $c$ designated by the attacker for inputs $\boldsymbol{x}$  implanted with a trigger chosen by the attacker.
The formal definition is shown as follows. \(y\) here means the correct label of the input $\boldsymbol{x}$.
\begin{equation}
\label{definition}
\mathcal{C}_\epsilon(\mathcal{F}_\theta(\boldsymbol{x}))=y, \quad \mathcal{C}_\epsilon(\mathcal{F}_\theta(\mathcal{I}_\phi(\boldsymbol{x})))=c    
\end{equation}


The attacker's goal is to modify a clean image encoder to create a backdoored version that meets two key objectives: \ding{172} \emph{Effectiveness}: The backdoored model should maintain a high attack success rate while preserving accuracy in benign conditions, keeping backdoored accuracy close to clean accuracy for downstream classifiers. \ding{173} \emph{Naturalness and Stealthiness}: The triggered sample should appear authentic and natural to avoid detection by human inspection. 

\noindent
\textbf{Attacker’s Knowledge and Capabilities.}\
Following BadEncoder~\cite{Jia_Liu_Gong_2022}, we assume that the attacker has access to a pre-trained clean image encoder and the attacker has full knowledge about the pre-trained encoder, such as the SSL method and the detailed contrastive augmentation operation used in pre-training.
Additionally, it is presumed that the attacker can access a collection of unlabeled images, referred to as \textit{shadow dataset}. The attacker is also assumed to have access to a few images from the Internet, called \textit{reference inputs}, for each combination of a target downstream task and a target class. We assume that the attacker can manipulate the training procedure to create an encoder with embedded backdoors. Accordingly, the attacker also has access to the augmentation transforms used to pre-train the encoder, which can be utilized in INACTIVE to generate stealthy and effective backdoor triggers.
However, we assume that the attacker cannot interfere with the training process of these downstream classifiers, such as the training dataset, model framework, and weights. Unlike data-poisoning-based methods such as CTRL~\cite{Li_2023_ICCV} and BLTO~\cite{sun2024backdoor}, our approach does not rely on matching distributions between pre-training and downstream datasets, allowing for broader applicability without interfering in downstream training data, models, or weights.



\subsection{Overarching Idea}
According to the previous observations, the key to enhancing the ASR in SSL is to disentangle the two overlapping distributions of the backdoor and augmentation transformation in the contrastive learning of the SSL's pre-training stage. 
Gray Scaling and Color Jittering are necessary augmentations used in the self-supervised learning and most of existing SSL methods (e.g., SimCLR~\cite{chen2020simple}, MoCo~\cite{he2020momentum}, SimSiam~\cite{chen2021exploring} and BYOL~\cite{grill2020bootstrap}) use them. A detailed summary of the augmentation operations in different mainstream SSL methods can be found in \autoref{tab:parameters}. 
Since these augmentations primarily alter the color semantics of inputs, HSV and HSL color spaces serve as ideal input spaces for capturing and enlarging these effects~\cite{jiang2023color}. We aim to identify a trigger that escapes the inherent augmentation space of self-supervised learning by increasing the distance between backdoored samples and non-backdoored samples within the HSV and HSL color spaces.
Since \emph{we already know the augmentation ways in the pre-training stage}, thus we design $\mathcal{L}_{\text {disentangle}}$ to quantify the distributional gap between images in the two batches, which involves measuring the difference in color characteristics.
To further enlarge the distributional gap, we design 
$\mathcal{L}_{\text {alignment}}$ to pull close the features of backdoor images and reference images.
Moreover, while we try to expand the distributional difference between the backdoor trigger and the augmentation transformation, an excessively large gap might result in a significant divergence between the backdoored image and the original one. This could, in turn, diminish the naturalness of the backdoored image and reduce the stealthiness of the trigger. Hence we design $\mathcal{L}_{\text {stealthy}}$ to blend the backdoor seamlessly with the original image. 

\subsection{Our Approach: INACTIVE}
\label{define}
In our context, we refer to a clean pre-trained image encoder and its backdoor-injected one as \( \mathcal{F}_\theta \) and \( \mathcal{F}_\theta' \). Given any pair of a downstream task and its corresponding target class, labeled as $\left(T_i, y_i\right)$, the attacker gathers a collection of reference inputs denoted by \( R_i = \{x_{i1}, x_{i2}, \ldots, x_{ir_i}\} \) from the specified target class \( y_i \), where \( r_i \) represents the number of reference inputs for \( (T_i, y_i) \), with $i=1,2, \cdots, t$. Moreover,
for each pair $\left(T_i, y_i\right)$, the attacker chooses a trigger $\boldsymbol{e}_i$ to implant into samples in the shadow dataset $\mathcal{D}_s$. We denote a clean input $\boldsymbol{x}$ embedded with a trigger as $\boldsymbol{x}'$, which is called a backdoored input.

\noindent
\textbf{Enhancing Distributional Gap Between Backdoor Images and Augmented Images.}\
To effectively enlarge the distributional distinction between the backdoor and the augmentation in SSL, we design $\mathcal{L}_{\text {disentangle}}$ for scenarios where augmentation transformations might weaken or obscure the pattern of backdoor triggers, leading to a decrease in the ASR. Utilizing $\mathcal{L}_{\text {disentangle}}$ ensures that the distinctiveness of the backdoor is maintained even in the face of various image transformations. 
The disentangle loss is defined as follows.
\begin{equation}
\label{disentangle}
\begin{array}{rl}
\mathcal{L}_{\text {disentangle}}=\\
 -\frac{1}{\left|\mathcal{D}_s\right|} \sum\limits_{\boldsymbol{x} \in D_s} & \|H\left(\boldsymbol{x} '\right)- H(\tilde{\boldsymbol{x}})\|_2+ \|S\left(\boldsymbol{x} '\right)-S(\tilde{\boldsymbol{x}})\|_2 \\
+ & \|V\left(\boldsymbol{x} '\right)-V(\tilde{\boldsymbol{x}})\|_2 +  \|L\left(\boldsymbol{x} '\right)- L(\tilde{\boldsymbol{x}})\|_2,
\end{array}
\end{equation}


where $H$,$S$,$V$,$L$ denote Hue, Saturation, Value, and Lightness from HSV and HSL color spaces. We denote an input $\boldsymbol{x}$ augmented by the transformations used in the encoder's pre-training stage as $\tilde{\boldsymbol{x}}$. $\left|\mathcal{D}_s\right|$ denotes the sample number in the shadow dataset.
$\|u-v\|_2$ denotes the $\ell_2$ distance between sample $u$ and sample $v$. 

\noindent
\textbf{Feature Alignment Between Backdoored and Reference Images.}\
Following BadEncoder,
we enhance the backdoor attack effectiveness by making the compromised image encoder outputs similar feature embeddings for any sample injected with backdoor $\boldsymbol{x} '$ in the shadow dataset $\mathcal{D}_s$ and the reference inputs $\mathcal{R}_i$ of a pair $\left(T_i, y_i\right)$. Consequently, a compromised downstream classifier developed from our compromised image encoder is inclined to assign identical labels to both reference samples $\mathcal{R}_i$ and to any compromised sample $\boldsymbol{x} '$. We call this process feature alignment between backdoored and reference images, and the $\mathcal{L}_{\text{alignment}}$ is defined as follows.
\begin{equation}
\label{alignment}
\mathcal{L}_{\text{alignment}}=-\frac{\sum_{i=1}^t \sum_{j=1}^{r_i} \sum_{\boldsymbol{x} \in \mathcal{D}_s} s\left(\mathcal{F}_\theta^{\prime}\left(\boldsymbol{x} '\right), \mathcal{F}_\theta^{\prime}\left(\boldsymbol{x}_{i j}\right)\right)}{\left|\mathcal{D}_s\right| \cdot \sum_{i=1}^t r_i},
\end{equation}
where \( s(\cdot,\cdot) \) is used to quantify the degree of similarity, for instance, cosine similarity, between a pair of feature embeddings. The term \( |D_s| \) denotes the count of samples within the shadow dataset, and the denominators serve the purpose of standardizing the losses.

\noindent
\textbf{Preserving Covert and Natural Backdoors with Expanded Distributional Gaps.}\
\label{Stealth}
We employ several metrics that measure the similarity between the backdoored image and the original one in both pixel and feature space to ensure that our trigger remains both natural and inconspicuous. To assess similarity in pixel space, we use SSIM and PSNR. Meanwhile, for high-level feature space comparisons, we first use LPIPS which better reflects the subjective experience of image quality and similarity.
Following \cite{10.1145/3319535.3354209} and \cite{tao2023distribution}, we also use Wasserstein distance~\cite{villani2009optimal} (WD) to reduce the distributional disparity between backdoored and clean samples.
$\mathcal{L}_{\text {stealthy }}$ is defined as follows:
\begin{equation}
\label{stealthy}
\begin{array}{rl}
\mathcal{L}_{\text {stealthy }}=\\\sum_{\boldsymbol{x} \in D_s} & 
\lambda_1 \cdot \mathrm{LPIPS}\left(\boldsymbol{x} ', \boldsymbol{x}\right)+ \mathrm{WD}\left(\mathcal{M}(\boldsymbol{x} '),\mathcal{M}(\boldsymbol{x})\right) \\ - & \lambda_2 \cdot \mathrm{PSNR}\left(\boldsymbol{x} ', \boldsymbol{x}\right) - \mathrm{SSIM}\left(\boldsymbol{x} ', \boldsymbol{x}\right),
\end{array}
\end{equation}
where $\lambda_1, \lambda_2$ are used to scale different loss terms to the same scale from 0 to 1.

\begin{algorithm}[t]
\scriptsize
\caption{Our backdoor attack INACTIVE}
\label{algo1}
 \hspace*{\algorithmicindent} \textbf{Input:} Pre-trained clean encoder \( \mathcal{F}_\theta^* \), shadow dataset \( \mathcal{D}_s \), reference input set \( \mathcal{R} \) \\
 \hspace*{\algorithmicindent} \textbf{Output:} Backdoored encoder \( \mathcal{F}_\theta \) , backdoor trigger injector \( \mathcal{I}_\phi \) 
\begin{algorithmic}[1]

\Function{Ours}{$\mathcal{F}_\theta^*$,  $\mathcal{D}_s$ ,  $\mathcal{R}$}
    \State \( \mathcal{F}_\theta \gets \mathcal{F}_\theta^* \); \( \hat{\mathcal{D}_s} \gets \) Augment samples in \( \mathcal{D}_s \); 
    \State \( \mathcal{I}_\phi \gets \) a pre-trained backdoor injector \( \mathcal{I}_\phi^* \) using  \autoref{algo2}
    \For{\( \text{iter} = 0 \) \textbf{to} \( \text{max\_epochs} \)}
        \State \( {\mathcal{D}_s}^{'} \gets \mathcal{I}_\phi(\mathcal{D}_s) \) 
        \State \( \mathcal{L}_{\text {disentangle}} \gets \) distribution difference between backdoor images $\boldsymbol{x}'$ and augmented images \( \tilde{\boldsymbol{x}} \), \(\forall \boldsymbol{x}' \in {\mathcal{D}_s}^{'}, \forall \tilde{\boldsymbol{x}} \in \hat{\mathcal{D}_s} \)         \Comment{ \autoref{disentangle} }

        \State \( \mathcal{L}_{\text {stealthy}} \gets \) distance between backdoor image $\boldsymbol{x}'$ and clean image \( \boldsymbol{x} \), \(\forall \boldsymbol{x}' \in {\mathcal{D}_s}^{'},\forall \boldsymbol{x} \in \mathcal{D}_s \) 
        \State \Comment{ \autoref{stealthy} }
        \State $
        \mathcal{L}_{\text{alignment}}=-\frac{\sum_{i=1}^t \sum_{j=1}^{r_i} \sum_{\boldsymbol{x} \in \mathcal{D}_s} s\left(\mathcal{F}_\theta\left(\boldsymbol{x} '\right), \mathcal{F}_\theta\left(\boldsymbol{x}_{i j}\right)\right)}{\left|\mathcal{D}_s\right| \cdot \sum_{i=1}^t r_i}$
        \Comment{ \autoref{alignment} }
        \State \( \mathcal{L}_{\text{injector}} = \mathcal{L}_{\text {stealthy}} + \alpha \cdot \mathcal{L}_{\text {disentangle}} + \beta \cdot \mathcal{L}_{\text {alignment}} \)
        \Comment{ \autoref{total loss} }
        \State \( \phi_\mathcal{I} = \phi_\mathcal{I} - lr_{1} \cdot \frac{\partial \mathcal{L}_{\text{injector}}}{\partial \phi_\mathcal{I}} \) 
        \State $\mathcal{L}_{\text {consistency }}=-\frac{\sum_{i=1}^t \sum_{j=1}^{r_i} s\left(\mathcal{F}_\theta^{\prime}\left(\boldsymbol{x}_{i j}\right), \mathcal{F}_\theta\left(\boldsymbol{x}_{i j}\right)\right)}{\sum_{i=1}^t r_i}$
        \Comment{\autoref{L_consistency}}
        \State $\mathcal{L}_{\text {utility}}=-\frac{1}{\left|\mathcal{D}_s\right|} \cdot \sum_{\boldsymbol{x} \in \mathcal{D}_s} s\left(\mathcal{F}_\theta^{\prime}(\boldsymbol{x}), \mathcal{F}_\theta(\boldsymbol{x})\right)$\Comment{\autoref{L_utility}}
        \State \( \mathcal{L}_{\text{encoder}} = \mathcal{L}_{\text {alignment}} + \mathcal{L}_{\text {consistency}} + \mathcal{L}_{\text {utility}} \)
        \State \( \theta_\mathcal{F} = \theta_\mathcal{F} - lr_{2} \cdot \frac{\partial \mathcal{L}_{\text{encoder}}}{\partial \theta_\mathcal{F}} \) 
    \EndFor
\EndFunction
\end{algorithmic}
\end{algorithm}

\begin{algorithm}[t]
\scriptsize
\caption{Pre-training backdoor injector}
\label{algo2}
 \hspace*{\algorithmicindent} \textbf{Input:} Shadow dataset \( \mathcal{D}_s \)
 \hspace*{\algorithmicindent} \textbf{Output:} Pre-trained backdoor injector \( \mathcal{I}_\phi \) 
\begin{algorithmic}[1]

\Function{Pre-training injector}{$\mathcal{D}_s$}
    \State \( \hat{\mathcal{D}_s} \gets \) Augment samples in \( \mathcal{D}_s \); \( \mathcal{I}_\phi \gets \text{Random initialization} \)
    \For{\( \text{iter} = 0 \) \textbf{to} \( \text{max\_epochs} \)}
        \State \( {\mathcal{D}_s}^{'} \gets \mathcal{I}_\phi(\mathcal{D}_s) \)  
        \State \( \mathcal{L}_{\text {disentangle}} \gets \) distribution difference between backdoor images $\boldsymbol{x}'$ and augmented images \( \tilde{\boldsymbol{x}} \), \(\forall \boldsymbol{x}' \in {\mathcal{D}_s}^{'}, \forall \tilde{\boldsymbol{x}} \in \hat{\mathcal{D}_s} \)     
        \Comment{ \autoref{disentangle} }
        \State \(\mathcal{L}_{\text {stealthy}} \gets \) distance between backdoor image $\boldsymbol{x}'$ and clean image \( \boldsymbol{x} \), \(\forall \boldsymbol{x}' \in {\mathcal{D}_s}^{'},\forall \boldsymbol{x} \in \mathcal{D}_s \) \Comment{\autoref{stealthy} }
        \State \( \mathcal{L}_{\text{ours}} = \mathcal{L}_{\text {stealthy}} + \mu \cdot \mathcal{L}_{\text {disentangle}} \)
        \State \( \phi_\mathcal{I} = \phi_\mathcal{I} - lr \cdot \frac{\partial \mathcal{L}_{\text{ours}}}{\partial \phi_\mathcal{I}} \) 
    \EndFor
\EndFunction
\end{algorithmic}

\end{algorithm}

\noindent
\textbf{Optimization Problem Formulation and Algorithm.}\
We have defined $\mathcal{L}_{\text {disentangle}}$, $\mathcal{L}_{\text {stealthy }}$, $\mathcal{L}_{\text{alignment}}$ in the sections above. Then we can define our INACTIVE as an optimization problem. Concretely, our backdoor trigger injector $\mathcal{I}_\phi$ is a solution to the subsequent optimization problem:
\begin{equation}
\label{total loss}
\min _{\theta_\mathcal{F}} \min _{\mathcal{I}_\phi} \mathcal{L}=\mathcal{L}_{\text {stealthy }}+\alpha \cdot \mathcal{L}_{\text {disentangle}}+\beta \cdot \mathcal{L}_{\text {alignment}},
\end{equation}
where $\alpha$ and $\beta$ serve as hyper-parameters to provide equilibrium among these three loss components. We adopt \autoref{algo1} to solve the optimization problem, where we alternatively optimize the backdoor injector and the compromised image encoder and output the final backdoored encoder \( \mathcal{F}_\theta \) and backdoor trigger injector \( \mathcal{I}_\phi \). Additionally, to speed up the optimization process and promote the backdoor attack efficacy, we adopt \autoref{algo2} to pre-train a backdoor injector to initialize the injector in \autoref{algo1}.
We use the U-Net architecture~\cite{ronneberger2015u} for the backdoor injector, as shown in the \autoref{unet}.


\section{Evaluation}
\label{Evaluation}
We first evaluate the effectiveness and stealthiness of INACTIVE using four datasets, followed by an assessment of its robustness against various backdoor defenses and noises. To demonstrate generalization, we conduct additional attacks on various SSL algorithms and a multi-modal model with different augmentations, detailed in \autoref{supple:SSL} and \autoref{sec:Multi-modal Models}. \autoref{supple:ablation_s} further validates each component's role. \autoref{supple:eval} examines parameter sensitivity and performance.

\begin{table*}[t]
\scriptsize
\setlength\tabcolsep{4pt}
\centering
\resizebox{\textwidth}{!}{
    \begin{tabular}{@{}cccccccccccccccccccc@{}}
    \toprule
    \multirow{2}{*}{\makecell{Pre-training \\ Dataset}} & \multirow{2}{*}{\makecell{Downstream \\ Dataset}} & No Attack & \multicolumn{2}{c}{\makecell{BadEncoder + \\WaNet trigger}} &  & \multicolumn{2}{c}{\makecell{BadEncoder + \\CTRL trigger}} &  & \multicolumn{2}{c}{\makecell{BadEncoder + \\Ins-Kelvin trigger}} &  & \multicolumn{2}{c}{\makecell{BadEncoder + \\Ins-Xpro2 trigger}} &  & \multicolumn{2}{c}{\makecell{DRUPE + \\Patch trigger}} &  & \multicolumn{2}{c}{Ours} \\ \cmidrule(lr){3-3} \cmidrule(lr){4-5} \cmidrule(lr){7-8} \cmidrule(lr){10-11} \cmidrule(lr){13-14} \cmidrule(lr){16-17} \cmidrule(l){19-20} 
     &  & CA & BA$\uparrow$ & ASR$\uparrow$ &  & BA$\uparrow$ & ASR$\uparrow$ &  & BA$\uparrow$ & ASR$\uparrow$ &  & BA$\uparrow$ & ASR$\uparrow$ &  & BA$\uparrow$ & ASR$\uparrow$&  & BA$\uparrow$ & ASR$\uparrow$ \\ \midrule
    \multirow{3}{*}{STL10} & CIFAR10 & 86.77 & 84.43 & 10.28 &  & 87.19 & 8.72 &  & 86.75 & 18.63 &  & 86.85 & 16.83 &  & 84.36 & 98.39 &  & 87.11 & \textbf{99.58} \\
     & GTSRB & 76.12 & 74.45 & 5.23 &  & 77.57 & 8.17 &  & 76.49 & 72.95 &  & 76.71 & 14.02 &  & 75.93 & 96.09 &  & 75.82 & \textbf{97.97} \\
     & SVHN & 55.35 & 58.29 & 16.83 &  & 54.29 & 3.32 &  & 56.67 & 38.03 &  & 58.42 & 18.68 &  & 75.64 & 96.68 &  & 58.62 & \textbf{99.76} \\ \midrule
    \multirow{3}{*}{CIFAR10} & STL10 & 76.14 & 72.73 & 9.78 &  & 75.73 & 16.85 &  & 74.89 & 1.16 &  & 74.11 & 5.91 &  & 74.43 & 96.72 &  & 74.02 & \textbf{99.68} \\
     & GTSRB & 81.84 & 75.85 & 5.46 &  & 79.94 & 97.95 &  & 78.56 & 2.50 &  & 75.08 & 42.40 &  & 80.35 & 97.22 &  & 79.15 & \textbf{98.73} \\
     & SVHN & 61.52 & 54.79 & 17.99 &  & 66.33 & 40.91 &  & 68.49 & 22.13 &  & 68.95 & 30.91 &  & 76.02 & 96.23 &  & 63.67 & \textbf{98.79} \\ \midrule
    Average & / & 72.96 & 70.09 & 10.93 &  & 73.51 & 29.32 &  & 73.64 & 25.90 &  & 73.35 & 21.46 &  & 77.79 & 96.89 &  & 73.10 & \textbf{99.09} \\ \bottomrule
    \end{tabular}%
}
\caption{
Effectiveness comparison to representative backdoor attacks in SSL with different triggers (CA(\%), BA(\%), and ASR(\%)).
We compare our method to BadEncoder~\cite{Jia_Liu_Gong_2022} with various existing stealthy triggers. We also include the results of DRUPE~\cite{tao2023distribution} with their default visible patch trigger.
We include CTRL here to demonstrate that it is ineffective across various downstream datasets.
\emph{Our approach constantly achieves the highest ASRs while maintaining the accuracy on clean samples of the downstream classifiers trained on the backdoored encoder.}}
\label{tab:comparison}
\vspace{-5pt}

\end{table*}

\subsection{Experimental Setup}

\noindent
\textbf{Datasets.}\
We utilize four image datasets, i.e. CIFAR10~\cite{krizhevsky2009learning}, STL10~\cite{pmlr-v15-coates11a}, GTSRB~\cite{STALLKAMP2012323}, SVHN~\cite{37648} and ImageNet~\cite{russakovsky2015imagenet} to evaluate our method, which are also frequently used in backdoor attacks research~\cite{Jia_Liu_Gong_2022,nguyen2021wanet}. More details are introduced in \autoref{supple:exp_detail}.

\noindent
\textbf{Evaluation Metrics.}\
To assess the effectiveness of our method, we employ three metrics following existing works~\cite{10.1145/3450569.3463560, Jia_Liu_Gong_2022}: \emph{Clean Accuracy (CA)}: the accuracy of a clean downstream classifier on clean testing images from the downstream dataset; \emph{Benign Accuracy (BA)}: the accuracy of a backdoored downstream classifier on the same clean testing images from the downstream dataset; \emph{Attack Success Rate (ASR)}: the success rate of backdoor attacks. 
To evaluate the stealthiness and naturalness of the backdoor triggers, we employ three metrics following existing works~\cite{jiang2023color}: \emph{SSIM}~\cite{ssim}, \emph{PSNR}~\cite{psnr}, \emph{LPIPS}~\cite{Zhang_2018_CVPR}, \emph{Feature Similarity Indexing Method (FSIM)}~\cite{5705575} and \emph{Fréchet Inception Distance(FID)}~\cite{NIPS2017_8a1d6947}. Higher SSIM, PSNR, FSIM and lower LIPIPS, FID indicate better stealthiness and naturalness of the generated backdoored images.


\noindent
\textbf{SSL Frameworks.}\
In the pre-training stage, we employ SimCLR~\cite{chen2020simple} by default to train a ResNet18~\cite{he2016deep} model, serving as our image encoder. Furthermore, we prove the effectiveness of our method on other SSL frameworks, i.e., MoCo~\cite{he2020momentum}, BYOL~\cite{grill2020bootstrap}, SimSiam~\cite{chen2021exploring}, SwAV~\cite{caron2020unsupervised}, and CLIP~\cite{radford2021learning} in \autoref{supple:SSL} and \autoref{sec:Multi-modal Models}.

\noindent
\textbf{Attack Baselines.}\
We select two Instagram filters, Kelvin and Xpro2, as baseline triggers for aesthetic enhancements~\cite{pilgram,liu2019abs}. Additionally, WaNet~\cite{nguyen2021wanet}, CTRL~\cite{Li_2023_ICCV}, and ISSBA~\cite{li2021invisible} are chosen for their stealthiness and high ASR. These triggers are injected into compromised encoders using BadEncoder. We also include DRUPE~\cite{tao2023distribution}, a SOTA backdoor method using SimCLR and a patch trigger, as a baseline. To ensure a fair comparison, we evaluate our method against CTRL~\cite{Li_2023_ICCV}, SSLBKD~\cite{Saha_Tejankar_Koohpayegani_Pirsiavash_2022}, POIENC~\cite{liu2022poisonedencoder}, and BLTO~\cite{sun2024backdoor} using the same CIFAR10 as the pre-trained and downstream dataset under SimCLR, BYOL, and SimSiam. For SSLBKD, the trigger is randomly placed, while for SSLBKD-fixed, it's in the lower-right corner. We show more 
 experimental settings and details in \autoref{supple:exp_detail}.

\begin{table}[t]
\scriptsize
\centering
\setlength\tabcolsep{2pt}
\begin{tabular}{@{}cccccccccc@{}}
\toprule
\multirow{4}{*}{Attack} & \multirow{4}{*}{Invisible} & \multicolumn{8}{c}{SSL Method} \\ \cmidrule(l){3-10} 
 &  & \multicolumn{2}{c}{SimCLR} &  & \multicolumn{2}{c}{BYOL} &  & \multicolumn{2}{c}{SimSiam} \\ \cmidrule(lr){3-4} \cmidrule(lr){6-7} \cmidrule(l){9-10} 
 &  & BA$\uparrow$ & ASR$\uparrow$ &  & BA$\uparrow$ & ASR$\uparrow$ &  & BA$\uparrow$ & ASR$\uparrow$ \\ \midrule
POIENC~\cite{liu2022poisonedencoder} & \ding{53} & 80.50 & 11.10 &  & 81.70 & 10.70 &  & 81.90 & 10.70 \\
SSLBKD~\cite{Saha_Tejankar_Koohpayegani_Pirsiavash_2022} & \ding{53} & 79.40 & 33.20 &  & 80.30 & 46.20 &  & 80.60 & 53.10 \\
SSLBKD (fixed)~\cite{Saha_Tejankar_Koohpayegani_Pirsiavash_2022} & \ding{53} & 80.00 & 10.50 &  & 82.30 & 11.20 &  & 81.90 & 10.70 \\
CTRL~\cite{Li_2023_ICCV} & \ding{51} & 80.50 & 85.30 &  & 82.20 & 61.90 &  & 82.00 & 74.90 \\
BLTO~\cite{sun2024backdoor} & \ding{53} & 90.10 & 91.27 &  & 91.21 & 94.78 &  & 90.18 & 84.63 \\
Ours & \ding{51} & 90.19 & \textbf{100.00} &  & 93.01 & \textbf{99.99} &  & 91.01 & \textbf{99.99} \\ \bottomrule
\end{tabular}
\caption{Effectiveness comparison to data-poisoning-based backdoor attacks in SSL with their default triggers. We show the results of BA(\%), and ASR(\%) with the same pre-trained and downstream dataset CIFAR10. Since data poisoning-based methods require matched distributions between pre-training and downstream, we use the same pre-trained and downstream datasets. Our threat model is different from theirs, and our method can be applied when the distributions of pre-training and downstream datasets are different. This table's key aim is to demonstrate that our method achieves much higher ASR than them.}
\label{tab:comparion2}
\vspace{-5pt}
\end{table}



\subsection{Effectiveness Evaluation}
\label{Effectiveness}
\noindent
\textbf{Effective Attack.}\
As shown in \autoref{tab:comparison}, with different pre-trained and downstream datasets, our method achieves a high average ASR of 99.09\% across various datasets. Additionally,  \autoref{tab:comparion2} demonstrates that with the same pre-trained and downstream datasets, our approach also achieves nearly 100\% ASRs.
Our method outperforms all baseline methods in all scenarios, highlighting its robustness and superior effectiveness in executing successful backdoor attacks.

\noindent
\textbf{Accuracy Preservation.}\
The downstream classifiers trained on the backdoored encoder maintain good accuracy on clean samples, as shown in  \autoref{tab:comparison}. 
The average BA is 73.10\% compared to the average CA of 72.96\%, with the difference within 1\%.
This suggests that the backdoor introduced by our method does not compromise the classifier's ability to label clean images correctly. 
This is because $\mathcal{L}_{\text {utility}}$ guarantees that the backdoored and clean image encoders yield similar feature vectors for clean inputs.

\subsection{Stealthiness Evaluation}
\label{Stealthiness}



\begin{table}[t]
\centering
\scriptsize
\setlength\tabcolsep{3pt}
\begin{tabular}{cccccc}
\toprule
Method & SSIM$\uparrow$ & PSNR$\uparrow$ & LPIPS$\downarrow$ & FSIM$\uparrow$ & FID$\downarrow$ \\
\midrule
Badencoder~\cite{Jia_Liu_Gong_2022}/DRUPE~\cite{tao2023distribution} & 0.8355 & 14.1110 & 0.07693 & 0.820 & 53.363 \\
CTRL~\cite{Li_2023_ICCV} & 0.9025 & 32.4098 & \textbf{0.00034} & 0.865 & 71.138 \\
WaNet~\cite{nguyen2021wanet} & 0.7704 & 14.2372 & 0.07432 & 0.662 & 98.092 \\
Ins-Kelvin~\cite{liu2019abs} & 0.4955 & 16.1925 & 0.14000 & 0.677 & 96.449 \\
Ins-Xpro2~\cite{liu2019abs} & 0.5981 & 17.9173 & 0.04434 & 0.817 & 35.084 \\
POIENC~\cite{liu2022poisonedencoder} & 0.1214 & 11.2787 & 0.15867 & 0.597 & 172.220 \\
SSLBKD~\cite{Saha_Tejankar_Koohpayegani_Pirsiavash_2022} & 0.8737 & 16.2414 & 0.09640 & 0.891 & 118.320 \\
BLTO~\cite{sun2024backdoor} & 0.8417 & 29.6756 & 0.00941 & 0.950 & 36.385 \\
Ours & \textbf{0.9633} & \textbf{35.8649} & 0.00896 & \textbf{0.969} & \textbf{16.320} \\
\bottomrule
\end{tabular}
\caption{Stealthiness comparison to existing methods on CIFAR10. Our method remains stealthy. Detailed data are shown in~\autoref{tab:stealthiness_detail}.}
\label{tab:stealthiness-evaluation}
\vspace{-5pt}
\end{table}

\noindent
\textbf{Algorithmic Metrics.}\
We first compare the average SSIM, PSNR, and LPIPS when the pre-trained dataset is CIFAR10 and downstream datasets are STL10, GTSRB, and SVHN injected with these backdoor triggers to compare the stealthiness of various backdoor attack methods.
 \autoref{tab:stealthiness-evaluation} indicates that our method exhibits strong stealthiness advantages with an average of 0.9633 SSIM, 35.8649 PSNR, 0.00896 LIPIS, 0.969 FSIM, and 16.320 FID indicating minimal structural changes to the images, hardly detectable noise, almost negligible perceptual difference between the original and perturbed images. 
Although CTRL achieves a better LPIPS, our method outperforms it in both SSIM and PSNR. Additionally, our average ASR is 99.09\%, significantly higher than CTRL 29.32\%(see~\autoref{tab:comparison}), indicating that our method is more effective overall.
More detailed data across various datasets, i.e., CIFAR10, STL10, GTSRB, SVHN, and ImageNet are shown in~\autoref{tab:stealthiness_detail} and \autoref{tab:stealthiness_imagenet}.

\subsection{Robustness Evaluation}
\label{robustness}
To assess the resilience of our method against current backdoor defenses, we deploy various SOTA backdoor defense strategies, i.e., DECREE~\cite{Feng_2023_CVPR}, Beatrix~\cite{ma2022beatrix}, ASSET~\cite{ASSET}, Neural Cleanse (NC)~\cite{Wang_Yao_Shan_Li_Viswanath_Zheng_Zhao_2019}, STRIP~\cite{gao2019strip}, Grad-CAM~\cite{Selvaraju_2017_ICCV} for evaluation. Additionally, to further test the robustness of our method, we evaluate its endurance against the following commonly studied noises, i.e., JPEG compression~\cite{10.1145/3219819.3219910,dong2019evading}, Poisson noise~\cite{Agarwal_Ratha_Singh_Vatsa_2023,Xu_Zhang_Zhang_2022}, and Salt\&Pepper noise~\cite{Li_Li_2020,Agarwal_Ratha_Singh_Vatsa_2023}.
We also design an adaptive defense method for INACTIVE.
\emph{We show that INACTIVE cannot be defended by STRIP, NC, Grad-CAM, noises, and adaptive defense in \autoref{supple:defense}.}

\noindent
\textbf{DECREE.}\
DECREE~\cite{Feng_2023_CVPR} identifies trojan attacks in pre-trained encoders by flagging an encoder as compromised if the reversed trigger’s $\mathcal{L}^1$ norm proportion falls below a 0.1 threshold. As shown in \autoref{tab:decree-results}, the $\mathcal{P} \mathcal{L}^1$-Norm for each pre-trained and downstream dataset pair exceeds this threshold, so DECREE fails to detect backdoored encoders created by INACTIVE. This is because our invisible trigger breaks DECREE’s assumption of a visible patch trigger, and our stealthy loss further narrows the distribution gap between backdoored and normal data, masking internal model anomalies.

\begin{table}[t]
\centering
\scriptsize
\setlength\tabcolsep{5pt}
\begin{tabular}{@{}ccc@{}}
\toprule
Pre-trained Dataset & Downstream Dataset & $\mathcal{P} \mathcal{L}^1$-Norm \\ 
\midrule
\multirow{3}{*}{CIFAR10} & STL10 & 0.25 \\
                         & SVHN  & 0.39 \\
                         & GTSRB & 0.15 \\ 
\midrule
\multirow{3}{*}{STL10}   & CIFAR10 & 0.21 \\
                         & SVHN    & 0.34 \\
                         & GTSRB   & 0.20 \\
\bottomrule
\end{tabular}
\caption{Evaluation results of DECREE \cite{Feng_2023_CVPR}. A model is judged as backdoored if its $\mathcal{P} \mathcal{L}^1$-Norm \textless 0.1.}
\label{tab:decree-results}
\vspace{-5pt}
\end{table}

\noindent
\textbf{Beatrix.}\
Beatrix~\cite{ma2022beatrix} identifies poisoned samples by detecting abnormalities in the feature space. We use two pretraining datasets, CIFAR-10 and STL-10, and create backdoored encoders using BadEncoder and INACTIVE. By sampling 500 clean inputs and 500 poisoned samples, we applied Beatrix to differentiate them. We find (see \autoref{tab:Beatrix}) that Beatrix effectively recognizes poisoned samples from BadEncoder with over 93\% accuracy. However, Beatrix struggles to identify poisoned samples from INACTIVE, with a detection accuracy of below 50\% on both CIFAR-10 and STL-10, which is like random guessing. 
We further analyze the reasons for the defense failure in \autoref{supple:defense}.
\begin{table}[t]
\centering
\scriptsize
\setlength\tabcolsep{4pt}
\begin{tabular}{@{}ccccccc@{}}
\toprule
Encoder & Method & TP & FP & FN & TN & Acc \\ 
\midrule
\multirow{2}{*}{CIFAR-10} & BadEncoder & 499 & 24 & 1 & 476 & 97.50\% \\
                          & Ours       & 0   & 24 & 500 & 476 & 47.60\% \\ 
\midrule
\multirow{2}{*}{STL-10}   & BadEncoder & 458 & 24 & 42 & 476 & 93.40\% \\
                          & Ours       & 5   & 24 & 495 & 476 & 48.10\% \\ 
\bottomrule
\end{tabular}
\caption{Detection results by Beatrix \cite{ma2022beatrix}. It struggles to detect poisoned samples from ours.}
\label{tab:Beatrix}
\vspace{-15pt}
\end{table}

\noindent
\textbf{ASSET.}\
ASSET aims to distinguish between backdoored and clean samples by eliciting distinct behaviors in the model when processing these two data types, facilitating their separation~\cite{ASSET}. We replicate their defensive techniques on our backdoored CIFAR-10 dataset. Specifically, we applied our synthesized trigger to CIFAR-10 (with a target label of 0) to create a poisoned version of CIFAR-10, maintaining a 100\% poisoning rate as our default setting. The feature extractor used is the ResNet18 backbone, trained on this poisoned CIFAR-10 dataset. 

The True Positive Rate (TPR) measures how effectively a backdoor detection method identifies backdoored samples, with a higher TPR (closer to 100\%) indicating stronger filtering capability. The False Positive Rate (FPR) reflects the precision of this filtering: when TPR is sufficiently high, FPR shows the trade-off, highlighting the proportion of clean samples incorrectly flagged as backdoored. A lower FPR suggests fewer clean samples are mistakenly discarded, ensuring more clean data is retained for further use. Based on ASSET's metrics, we calculated the TPR as 7.14\% and the FPR as 1.8\%, indicating that our poisoned data can largely evade ASSET's detection.

\subsection{Generalization to Large-scale Dataset}
\label{real}

We assess the generalization of our method on a large-scale dataset by attacking an ImageNet-pre-trained encoder from Google~\cite{chen2020simple}. We compare our method’s performance with ISSBA, which is also trained and tested on ImageNet in its paper. Experimental setups are detailed in \autoref{supple:exp_detail}.

\noindent
\textbf{Experimental Results.}\
 \autoref{tab:performance-metrics-imagenet} indicates that \emph{our method is highly effective on ImageNet}, with an average 98.66\% ASR across different datasets. 
Moreover,  \autoref{tab:stealthiness-evaluation-imagenet}
indicates the high SSIM and PSNR values and low LPIPS values, demonstrating that \emph{the perturbations made by INACTIVE are almost imperceptible.} 
Moreover, the average 83.91\% BA is close to the average 83.59\% CA, indicating \emph{our attack maintains accuracy for the given downstream task despite the backdoor.} 
Additionally, \emph{both our ASR and BA are much higher than those of the baseline ISSBA}~\cite{li2021invisible}, proving ours has better performance.

\begin{table}[t]
\centering
\scriptsize
\setlength\tabcolsep{5pt}
\begin{tabular}{ccccccc}
\toprule
\multirow{2}{*}{Downstream Dataset}  & No Attack  & \multicolumn{2}{c}{ISSBA~\cite{li2021invisible}} & \multicolumn{2}{c}{Ours} \\ 
\cmidrule(lr){2-2} \cmidrule(lr){3-4} \cmidrule(lr){5-6}
& CA & BA$\uparrow$ & ASR$\uparrow$ & BA$\uparrow$ & ASR$\uparrow$ \\ 
\midrule
STL10 & 95.68 & 92.58 & 9.97 & 93.48 & 100.00 \\
GTSRB & 80.32 & 66.29 & 5.10 & 82.84 & 96.00 \\
SVHN  & 74.77 & 67.67 & 18.03 & 75.40 & 99.99 \\
\midrule
Average  & 83.59 & 75.51  & 11.03 & 83.91 & 98.66 \\

\bottomrule
\end{tabular}
\caption{Comparative results (CA(\%), BA(\%), and ASR(\%)) of ISSBA~\cite{li2021invisible} and our attack on \emph{ImageNet}. Ours constantly achieves the highest ASRs while maintaining accuracy on clean samples of the downstream classifiers.}
\label{tab:performance-metrics-imagenet}
\vspace{-5pt}
\end{table}

\begin{table}[]
\centering
\scriptsize
\setlength\tabcolsep{4pt}
\begin{tabular}{cccc}
\toprule
Method & Average SSIM$\uparrow$ & Average PSNR (dB)$\uparrow$ & Average LPIPS$\downarrow$ \\
\midrule
ISSBA~\cite{li2021invisible} & 0.7329 & 31.3496 & 0.12424 \\
Ours & \textbf{0.9867} & \textbf{34.5733} & \textbf{0.01233} \\
\bottomrule
\end{tabular}

\caption{Stealthiness comparison on ImageNet.}
\label{tab:stealthiness-evaluation-imagenet}
\vspace{-5pt}

\end{table}

\section{Conclusions and Future Work}

In this paper, we propose an imperceptible and effective backdoor attack against self-supervised models based on the optimized triggers that are disentangled in the augmented transformation in the SSL.
Based on the evaluation across five different datasets and six SSL algorithms, our attack is demonstrated to be both highly effective and stealthy. It also effectively bypasses existing backdoor defenses. For future work, it would be beneficial to expand the scope of research to include various other domains of machine learning, such as NLP and audio processing.


{
    \small
    \bibliographystyle{ieeenat_fullname}
    \bibliography{main}
}

\clearpage
\setcounter{page}{1}
\maketitlesupplementary

\setcounter{figure}{0}
\renewcommand{\thefigure}{A\arabic{figure}}
\setcounter{table}{0}
\renewcommand{\thetable}{A\arabic{table}}
\setcounter{section}{0}
\renewcommand{\thesection}{A\arabic{section}}
\setcounter{subsection}{0}
\renewcommand{\thesubsection}{A\arabic{section}.\arabic{subsection}}
\setcounter{equation}{0}
\renewcommand{\theequation}{A\arabic{equation}}
\setcounter{algorithm}{0}
\renewcommand{\thealgorithm}{A\arabic{algorithm}}

\section{Generalization to Other SSL Frameworks}
\label{supple:SSL}
SSL frameworks for Contrastive Learning can be normally based on the following categories: negative examples, self-distillation, and clustering. 
To further evaluate the efficacy of INACTIVE, we test our method across the representative SSL algorithms in each category apart from the default SimCLR, i.e., negative examples (MoCo~\cite{he2020momentum}), self-distillation (BYOL~\cite{grill2020bootstrap}, SimSiam~\cite{chen2021exploring}), and clustering (SwAV~\cite{caron2020unsupervised}). \autoref{tab:parameters} illustrates the detailed parameters for different augmentation transforms used in the SSL algorithms.
Other experiment settings are the same as \autoref{supple:exp_detail}.

\noindent
\textbf{Experimental Results.}\
\autoref{tab:SSLs} shows the evaluation results of applying INACTIVE to the encoders pre-trained utilizing various SSL algorithms on CIFAR-10. The SSL algorithms demonstrate high effectiveness in achieving successful backdoor attacks with ASR nearing or reaching perfection in many cases, with an average ASR of 99.62\%. Additionally, the algorithms excel in maintaining the stealthiness of these attacks, evidenced by a high average SSIM of 0.95 and a high average PSNR value of 28.42, alongside a low LPIPS value of 0.0086. These metrics across the SSL algorithms collectively indicate that while INACTIVE is highly effective at manipulating model behaviors, they do so in a way that remains largely undetectable through standard image quality assessments. 

\begin{table}[ht]
\centering
\scriptsize
\setlength\tabcolsep{4pt}
\begin{tabular}{@{}ccccccc@{}}
\toprule
\makecell{SSL\\ framework} & \makecell{Downstream\\ dataset} & BA(\%)$\uparrow$ & ASR(\%)$\uparrow$ & SSIM$\uparrow$ & PSNR$\uparrow$ & LPIPIS$\downarrow$ \\ \midrule
\multirow{3}{*}{MOCO~\cite{he2020momentum}} & STL10 & 72.56 & 99.75 & 0.95 & 26.74 & 0.0080 \\
 & GTSRB & 68.11 & 99.98 & 0.95 & 26.84 & 0.0070 \\
 & SVHN & 74.58 & 99.97 & 0.96 & 31.86 & 0.0030 \\ \midrule
\multirow{3}{*}{SimSiam~\cite{chen2021exploring}} & STL10 & 64.79 & 99.38 & 0.96 & 32.46 & 0.0019 \\
 & GTSRB & 28.99 & 99.90 & 0.96 & 24.26 & 0.0190 \\
 & SVHN & 52.24 & 100.00 & 0.96 & 26.04 & 0.0141 \\ \midrule
\multirow{3}{*}{BYOL~\cite{grill2020bootstrap}} & STL10 & 79.49 & 99.98 & 0.96 & 31.44 & 0.0025 \\
 & GTSRB & 70.70 & 100.00 & 0.96 & 35.34 & 0.0011 \\
 & SVHN & 42.76 & 99.74 & 0.96 & 32.12 & 0.0024 \\ \midrule
\multirow{3}{*}{SwAV~\cite{caron2020unsupervised}} & STL10 & 66.28 & 99.76 & 0.89 & 23.43 & 0.0258 \\
 & GTSRB & 79.39 & 100.00 & 0.97 & 25.42 & 0.0089 \\
 & SVHN & 76.71 & 96.94 & 0.96 & 25.05 & 0.0096 \\ \midrule
\multicolumn{2}{c}{Average} & 64.72 & 99.62 & 0.95 & 28.42 & 0.0086 \\ \bottomrule
\end{tabular}%
\caption{Evaluations on other SSL algorithms, which are used to pre-train the encoders on CIFAR-10. INACTIVE keeps highly effective and stealthy across these algorithms.}
\label{tab:SSLs}
\vspace{-10pt}
\end{table}

\begin{table}[t]
\centering
\scriptsize
\setlength\tabcolsep{3pt}
\begin{tabular}{@{}ccccccc@{}}
\toprule
Methods & CA(\%) & BA(\%)$\uparrow$ & ASR(\%)$\uparrow$ & SSIM$\uparrow$ & PSNR$\uparrow$ & LPIPIS$\downarrow$ \\ \midrule
Ours & \multirow{2}{*}{70.6} & 68.98 & 99.95 & 0.98 & 30.11 & 0.0179 \\
BadEncoder &  & 70.27 & 99.99 & 0.28 & 9.98 & 0.6105 \\ \bottomrule
\end{tabular}%

\caption{Attack results on the multi-shot classification of CLIP. INACTIVE attains comparable high BA and ASR to those of BadEncoder~\cite{Jia_Liu_Gong_2022}, yet it offers significantly enhanced stealthiness.}
\label{tab:CLIP}
\vspace{-10pt}
\end{table}

\section{Generalization to Multi-modal Models}
\label{sec:Multi-modal Models}
We utilize our INACTIVE to integrate a backdoor into the image encoder of the multi-modal model CLIP~\cite{radford2021learning}, which is comprised of both an image and a text encoder, pre-trained on 400 million internet-sourced (image, text) pairs. This backdoored image encoder is then employed to construct a multi-shot downstream classifier.


\noindent
\textbf{Experimental Results.}\
\autoref{tab:CLIP} presents the attack results of multi-shot classification using CLIP.
INACTIVE achieves a similar high BA and ASR as BadEncoder~\cite{Jia_Liu_Gong_2022} does, but it maintains much more stealthiness with high SSIM\&PSNR and low LPIPIS values than BadEncoder which uses a white patch as the backdoor trigger.

\section{Ablation Study}
\label{supple:ablation_s}
We utilize $\mathcal{L}_{\text{disentangle}}$ to enlarge the gap between the backdoored image distribution and the augmented image distribution. We use $\mathcal{L}_{\text{alignment}}$ to promote feature alignment between backdoored images and reference images. We use $\mathcal{L}_{\text{stealthy}}$ to make the backdoor trigger stealthy and natural. In \autoref{algo1}, we initialize a backdoor injector using a pre-trained one by \autoref{algo2}. 
We use STL10 as the dataset for pre-training encoders. 
In this section, we demonstrate the effects of each important component by ablating them respectively. 

As reported in \autoref{ablation}, the ASRs of our method decrease across all target downstream datasets when ablating each component respectively. As reported in \autoref{ablation2}, the average SSIM and PSNR drop dramatically and LPIPS increases a lot across all datasets when ablating $\mathcal{L}_{\text{stealthy}}$.
For example, in the ablation case of the target downstream dataset CIFAR10, the ASR decreases from 99.58\% to 67.84\%, 10.22\%, 75.27\% when ablating $\mathcal{L}_{\text{disentangle}}$, $\mathcal{L}_{\text{alignment}}$ and backdoor injector initialization respectively. This distinctly shows that the absence of key components significantly compromises the effectiveness of our backdoor attack methodology. 
Moreover, when ablating $\mathcal{L}_{\text{stealthy}}$, the SSIM drops from 0.99 to 0.1. The PSNR drops from 41.80 to 5.65. The LPIPS increases from 0.0002 to 0.55 indicating that the backdoored images without the $\mathcal{L}_{\text{stealthy}}$ are more easily detectable and differ a lot from the original images.

\begin{table}[t]
\scriptsize
\centering
\setlength\tabcolsep{1pt}

\begin{tabular}{@{}cccccccccccc@{}}
\toprule
\multirow{4}{*}{\makecell{Pre-training \\ Backdoor\\ injector}} & \multirow{4}{*}{$\mathcal{L}_{\text {disentangle}}$} & \multirow{4}{*}{$\mathcal{L}_{\text {alignment}}$} & \multirow{4}{*}{$\mathcal{L}_{\text {stealthy }}$} & \multicolumn{8}{c}{Target Downstream Dataset} \\ \cmidrule(l){5-12} 
 &  &  &  & \multicolumn{2}{c}{\multirow{1}{*}{CIFAR10}} &  & \multicolumn{2}{c}{\multirow{1}{*}{GTSRB}} &  & \multicolumn{2}{c}{\multirow{1}{*}{SVHN}} \\ \cmidrule(lr){5-6} \cmidrule(lr){8-9} \cmidrule(l){11-12} 
 &  &  &  & BA$\uparrow$ & ASR$\uparrow$ &  & BA$\uparrow$ & ASR$\uparrow$ &  & BA$\uparrow$ & ASR$\uparrow$ \\ \midrule
\multicolumn{1}{l}{} & \checkmark & \checkmark & \checkmark & 84.66 & 75.27 &  & 78.72 & 73.33 &  & 58.38 & 13.87 \\
\multicolumn{1}{l}{} &  & \checkmark & \checkmark & 85.71 & 67.84 &  & 74.05 & 52.26 &  & 56.41 & 62.56 \\
\checkmark & \checkmark &  & \checkmark & 87.20 & 10.22 &  & 65.49 & 7.46 &  & 56.01 & 23.44 \\
\checkmark & \checkmark & \checkmark & \checkmark & 87.11 & 99.58 &  & 75.82 & 97.97 &  & 58.62 & 99.76 \\ \bottomrule
\end{tabular}%
\caption{ASR(\%), BA(\%) in ablation studies. Ablating the $\mathcal{L}_{\text {alignment}}$ means ablating the one in $\mathcal{L}_{\text {injector}}$ instead of $\mathcal{L}_{\text{encoder}}$, and ablating the pre-trained backdoor injector means randomly initializing it. 
}
\label{ablation}
\end{table}

\begin{table}[t]
\centering
\setlength\tabcolsep{1pt}
\scriptsize
\begin{tabular}{@{}cccccccccccc@{}}
\toprule
\multirow{4}{*}{Method} & \multicolumn{11}{c}{Target Downstream Dataset} \\ \cmidrule(l){2-12} 
 & \multicolumn{3}{c}{CIFAR10} &  & \multicolumn{3}{c}{GTSRB} &  & \multicolumn{3}{c}{SVHN} \\ \cmidrule(lr){2-4} \cmidrule(lr){6-8} \cmidrule(l){10-12} 
 & SSIM$\uparrow$ & PSNR$\uparrow$ & LPIPS$\downarrow$ &  & SSIM$\uparrow$ & PSNR$\uparrow$ & LPIPS$\downarrow$  &  & SSIM$\uparrow$ & PSNR$\uparrow$ & LPIPS$\downarrow$  \\ \midrule
\makecell{w/o $\mathcal{L}_{\text{stealthy}}$} & 0.10 & 5.65 & 0.5500 &  & 0.02 & 5.20 & 0.4400 &  & 0.10 & 4.85 & 0.52 \\
Ours & 0.99 & 41.80 & 0.0002 &  & 0.99 & 43.58 & 0.0001 &  & 0.96 & 25.65 & 0.01 \\ \bottomrule
\end{tabular}
\caption{Results of ablating $\mathcal{L}_{\text{stealthy}}$. Images backdoored without the \(\mathcal{L}_{\text{stealthy}}\) are more readily identifiable and exhibit significant deviations from the original images.}
\label{ablation2}
\end{table}

\begin{table*}[t]
\centering
\scriptsize

\begin{tabular}{@{}cccccccl@{}}
\toprule
Parameter & SimCLR/CLIP & MOCO & SimSiam & BYOL & SwAV & Description \\ \midrule
input\_size & 32/224 & 32 & 32 & 32 & / & Size of the input image in pixels. \\
cj\_prob & 0.8 & 0.8 & 0.8 & 0.8 & 0.8 & Probability that color jitter is applied. \\
cj\_strength & 1 & 0.5 & 1 & 1 & 0.5 & \makecell[l]{Strength of the color jitter.\\ cj\_bright, cj\_contrast,  cj\_sat,  and cj\_hue \\ are multiplied by this value.} \\
cj\_bright & 0.4 & 0.4 & 0.4 & 0.4 & 0.8 & How much to jitter brightness. \\
cj\_contrast & 0.4 & 0.4 & 0.4 & 0.4 & 0.8 & How much to jitter contrast. \\
cj\_sat & 0.4 & 0.4 & 0.4 & 0.2 & 0.8 & How much to jitter saturation. \\
cj\_hue & 0.1 & 0.1 & 0.1 & 0.1 & 0.2 & How much to jitter hue. \\
min\_scale & / & 0.2 & 0.2 & 0.08 & / & \makecell[l]{Minimum size of the randomized crop\\ relative to the input\_size.} \\
random\_gray\_scale & 0.2 & 0.2 & 0.2 & 0.2 & 0.2 & Probability of conversion to grayscale. \\
vf\_prob & 0 & 0 & 0 & 0 & 0 & Probability that vertical flip is applied. \\
hf\_prob & 0.5 & 0.5 & 0.5 & 0.5 & 0.5 & Probability that horizontal flip is applied. \\
rr\_prob & 0 & 0 & 0 & 0 & 0 & Probability that random rotation is applied. \\
crop\_sizes & / & / & / & / & {[}32,32{]} & \makecell[l]{Size of the input image in\\ pixels for each crop category.} \\
crop\_counts & / & / & / & / & {[}2,2{]} & Number of crops for each crop category. \\
crop\_min\_scales & / & / & / & / & {[}0.14,0.14{]} & Min scales for each crop category. \\
crop\_max\_scales & / & / & / & / & {[}32,32{]} & Max scales for each crop category. \\
\bottomrule
\end{tabular}%
\caption{Parameters of Augmentation Transforms in SSL Algorithms.}
\label{tab:parameters}
\end{table*}


\section{Proof}
\label{sec:proof}

We aim to prove the following statement:

\noindent
\textbf{Theorem 3.1.} \textit{Given a perfectly-trained encoder $\mathcal{F}_\theta$ based on the augmentations sampled from the predefined augmentation space $S_A$, it is impossible to inject a backdoor with a trigger function $\mathcal{T} \in S_A$.}

\begin{proof}

The encoder $\mathcal{F}_\theta$ is said to be perfectly-trained if it achieves maximal similarity on any pair of augmented versions of the same training sample $x$, for all augmentations in the augmentation space $S_A$. Formally, this means:
\[
s(\mathcal{F}_\theta(A_1(x)), \mathcal{F}_\theta(A_2(x))) = 1, \quad \forall x \in \mathcal{X}, \, \forall A_1, A_2 \in S_A,
\]
where $s(\cdot, \cdot)$ denotes the similarity measure, which achieves its maximum value when the two inputs are identical in the feature space.
As a result, for all $x \in \mathcal{X}$ and any $A_1, A_2 \in S_A$, the representations produced by $\mathcal{F}_\theta$ are identical:
\[
\mathcal{F}_\theta(A_1(x)) = \mathcal{F}_\theta(A_2(x)).
\]

A backdoor injection introduces a trigger function $\mathcal{T}$ such that for any input $x$, the representation produced by $\mathcal{F}_\theta$ after applying the trigger, $\mathcal{F}_\theta(\mathcal{T}(x))$, maps to a target representation that is distinct from those of legitimate augmentations. 
In this case, we assume $\mathcal{T} \in S_A$, meaning the trigger function is an augmentation within the predefined augmentation space.
Since $\mathcal{T} \in S_A$, by the definition of the perfectly-trained encoder, the following holds for all $x \in \mathcal{X}$ and any $A_1, A_2 \in S_A$:
\[
\mathcal{F}_\theta(A_1(x)) = \mathcal{F}_\theta(\mathcal{T}(x)) = \mathcal{F}_\theta(A_2(x)).
\]

This implies that the feature representation of the input $x$ after applying the trigger $\mathcal{T}$ is indistinguishable from those produced by any other augmentations in $S_A$. As such, the trigger $\mathcal{T}$ fails to produce a unique or distinct representation in the feature space.
For a backdoor to be successfully injected, the trigger function $\mathcal{T}$ must produce a representation $\mathcal{F}_\theta(\mathcal{T}(x))$ that is distinct from those produced by other augmentations in $S_A$. However, this contradicts the property of the perfectly-trained encoder:
\[
\mathcal{F}_\theta(A_1(x)) = \mathcal{F}_\theta(\mathcal{T}(x)), \quad \forall A_1 \in S_A.
\]

Thus, $\mathcal{T}$ cannot create a distinct feature representation and, therefore, cannot function as a backdoor.
We have shown that for a perfectly-trained encoder $\mathcal{F}_\theta$, trained on augmentations from $S_A$, any trigger function $\mathcal{T} \in S_A$ fails to produce a distinct representation necessary for a backdoor injection. Therefore, it is impossible to inject a backdoor with a trigger function $\mathcal{T} \in S_A$.

\end{proof}

\section{More Evaluations}
\label{supple:eval}
\subsection{More Robustness Tests}
\label{supple:defense}

\begin{figure*}
    \centering
    \includegraphics[width=0.8\linewidth]{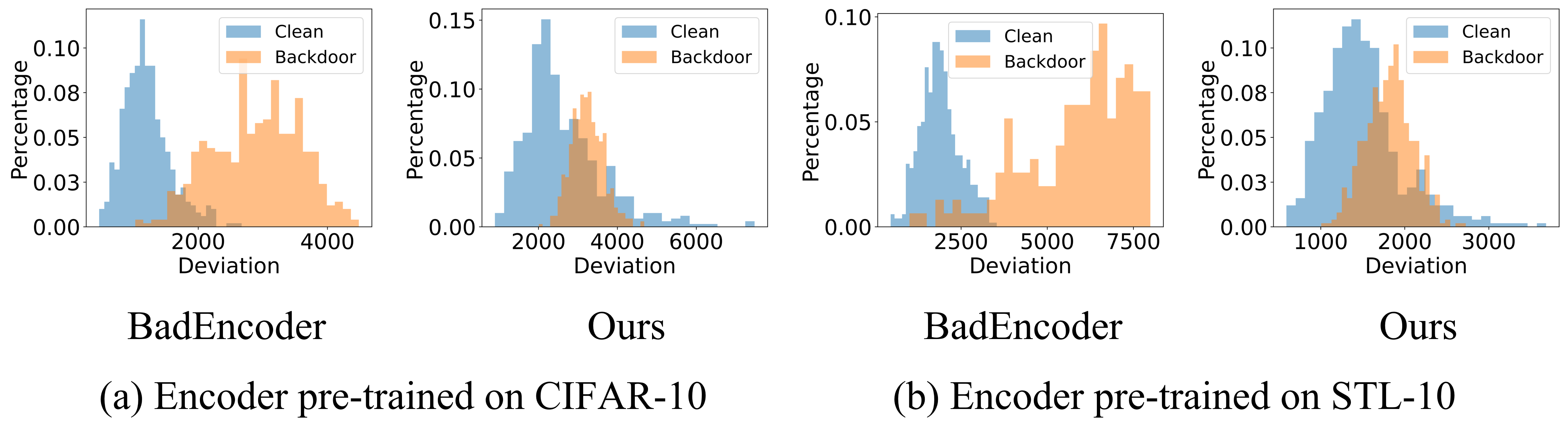}
    \caption{Beatrix's deviation distribution~\cite{ma2022beatrix} for clean and backdoored data, where blue bars represent clean samples and orange bars indicate poisoned ones. Poisoned samples from BadEncoder display markedly higher deviations than clean samples, allowing Beatrix to detect them effectively. However, the deviation distributions for clean inputs and poisoned samples from INACTIVE significantly overlap, complicating their differentiation.}
    \label{fig:Deviation}
\end{figure*}

\noindent
\textbf{Beatrix.}\
\label{supple:Beatrix}
Further examination of Beatrix's detection performance reveals that it determines whether a sample is poisoned based on the deviation in the feature matrix within the embedding space~\cite{ma2022beatrix}. A sample is deemed poisoned if the deviation exceeds a specific threshold. \autoref{fig:Deviation} displays the distribution of deviation values. Notably, poisoned samples by BadEncoder show significantly higher deviation than clean samples, enabling Beatrix to achieve high detection accuracy. Conversely, the deviation distributions between clean inputs and poisoned samples by INACTIVE overlap considerably, making them difficult to differentiate.

\noindent
\textbf{Neural Cleanse.}\
We then test the resilience of our method against a well-known reverse
engineering defense, namely Neural Cleanse (NC)~\cite{Wang_Yao_Shan_Li_Viswanath_Zheng_Zhao_2019}. 
Specifically, it utilizes the anomaly index to measure the deviation of the reconstructed triggers by their sizes, labeling models with an anomaly index exceeding two as Trojan-infected. Since NC is specifically designed for classifiers and cannot be directly used on image encoders, NC is employed to identify backdoors in a compromised downstream classifier. Moreover, the testing dataset from a targeted downstream dataset serves as the clean dataset. Experimental outcomes, as illustrated in \autoref{tab:neural-cleansing-results}, indicate that NC is unable to detect the Trojan model created by our approach. While effective in identifying patch-based Trojans~\cite{liu2018fine}, this method presumes a uniform trigger pattern at the pixel level across different samples. Our method circumvents this by employing input-dependent triggers, meaning the pixel-level Trojan alterations vary across samples. 

\begin{table}[t]
\centering
\scriptsize
\begin{tabular}{@{}ccc@{}}
\toprule
\textbf{{Pre-trained Dataset}} & \textbf{{Downstream Dataset}} & \textbf{{\makecell{Anomaly Index \\~\cite{Wang_Yao_Shan_Li_Viswanath_Zheng_Zhao_2019}}}} \\ 
\midrule
\multirow{3}{*}{CIFAR10} & STL10 & 1.17 \\
                         & SVHN  & 1.42 \\
                         & GTSRB & 1.61 \\ 
\midrule
\multirow{3}{*}{STL10}   & CIFAR10 & 0.88 \\
                         & SVHN    & 1.23 \\
                         & GTSRB   & 1.43 \\
\bottomrule
\end{tabular}
\caption{Evaluation results of neural cleanse \cite{Wang_Yao_Shan_Li_Viswanath_Zheng_Zhao_2019}. A model is judged as backdoored if its anomaly index \textgreater 2, so \textbf{our attack cannot be detected by it}.}
\label{tab:neural-cleansing-results}
\end{table}

\begin{figure*}[t]
    \centering
    \includegraphics[width=1\textwidth]{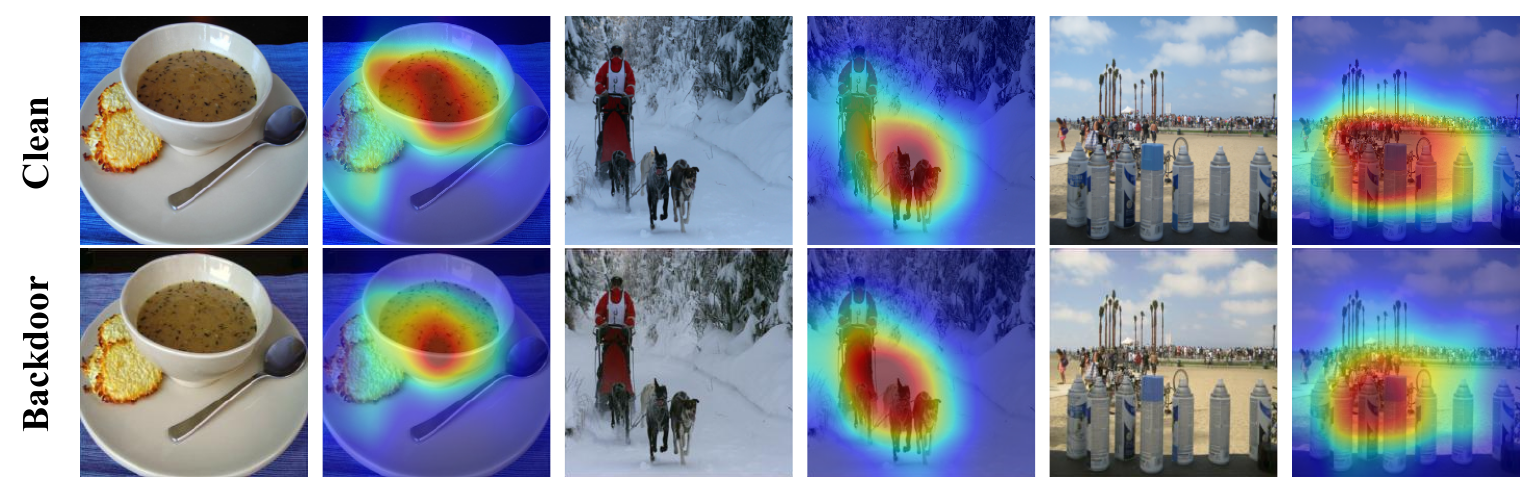}
    \caption{Resilience to Grad-CAM~\cite{Selvaraju_2017_ICCV}. The resemblance observed in these heatmaps indicates that INACTIVE is capable of resisting defenses that rely on Grad-CAM.}
    \label{gradcam}

\end{figure*}

\noindent
\textbf{Grad-CAM.}\
We assess INACTIVE's resilience to Grad-CAM~\cite{Selvaraju_2017_ICCV}, which generates a heatmap for a given model and input sample, where the heat intensity of each pixel reflects its significance in the model's final prediction. Grad-CAM is effective for identifying smaller Trojans~\cite{liu2018trojaning, gu2017badnets}, as these Trojans tend to produce high heat values in compact trigger zones, leading to abnormal heatmaps. Nevertheless, our backdoor transformation function alters the entire image, rendering Grad-CAM ineffective in detecting it. \autoref{gradcam} illustrates the comparison of heatmaps from both clean images and backdoored images created by our method. The similarity in these heatmaps suggests that INACTIVE can withstand defenses based on Grad-CAM.

\noindent
\textbf{Various Noises.}\
Next, we are going to introduce the noises, i.e., JPEG compression, Salt\&Pepper noise, and Poisson noise, applied in the robustness test.

\begin{itemize}
    \item JPEG Compression: JPEG is a prevalent image format. It is a common case that images undergo compression, especially during web transmission. The quality scale ranges from 1 to 100, with 100 being the highest quality and 1 being the lowest quality.
    \item Poisson Noise: It arises from the statistical nature of photon reception by image sensors, resulting in random noise of varying intensities. It is set between 1 and 10 to model different lighting conditions and sensor sensitivities.
    \item Salt\&Pepper Noise: This noise model mimics random pixel corruption due to image sensor errors, transmission faults, or system failures, characterized by randomly scattered white (salt) and black (pepper) pixels. It typically ranges from 0 to 1, but practical applications often use a much smaller range (e.g., 0.01 to 0.2) to avoid overwhelming the image with noise.
\end{itemize}

\begin{figure*}[t]
    \centering
    \includegraphics[width=1\textwidth]{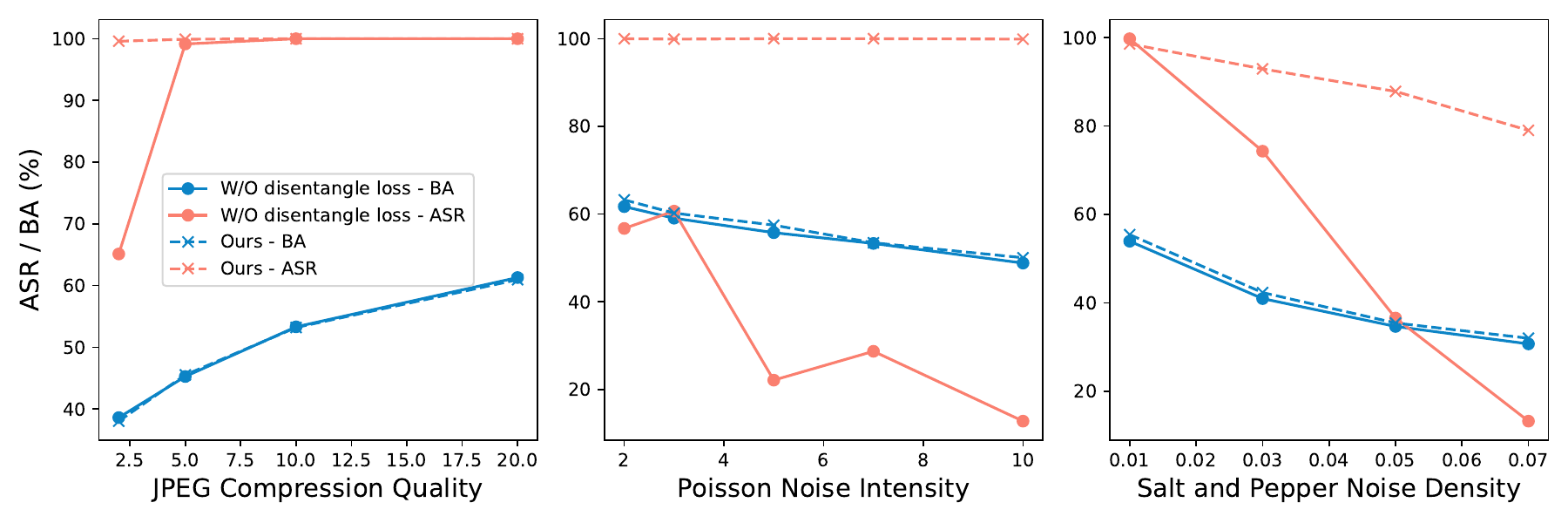}
    \caption{Resilience to JPEG compression, Salt\&Pepper noise, Poisson noise of different densities.}
    \label{robust}
\end{figure*}

Evaluating the robustness of our method across these noises helps ensure its performance is reliable even when image quality is compromised when encountering real-world image issues. We illustrate the evaluation results in \autoref{robust}, where the pre-trained dataset is CIFAR10 and the downstream dataset is STL10. The results prove that our method, which utilizes $\mathcal{L}_{\text {disentangle}}$, maintains high ASRs in most cases, and the ASR is significantly higher than the cases ablating $\mathcal{L}_{\text {disentangle}}$ in our method. For example, under different intensities of Poisson noise, the ASR of our method maintains nearly 100\%, while the ASR of the cases ablating $\mathcal{L}_{\text {disentangle}}$ in our method is unstable, ranging from 12\% to 61\%.
The outcome indicates that the $\mathcal{L}_{\text {disentangle}}$ plays a prominent role in promoting the robustness of our method. The reason may be that the $\mathcal{L}_{\text {disentangle}}$ encourages the model to learn features that remain invariant under the transformations. Namely, the model can still recognize the embedded triggers even when various noises or other image perturbations are applied. This aspect of feature preservation is key to ensuring that the backdoor attack remains effective regardless of the image transformations or certain noises applied.

\begin{figure*}[t]
    \centering
    \includegraphics[width=1\textwidth]{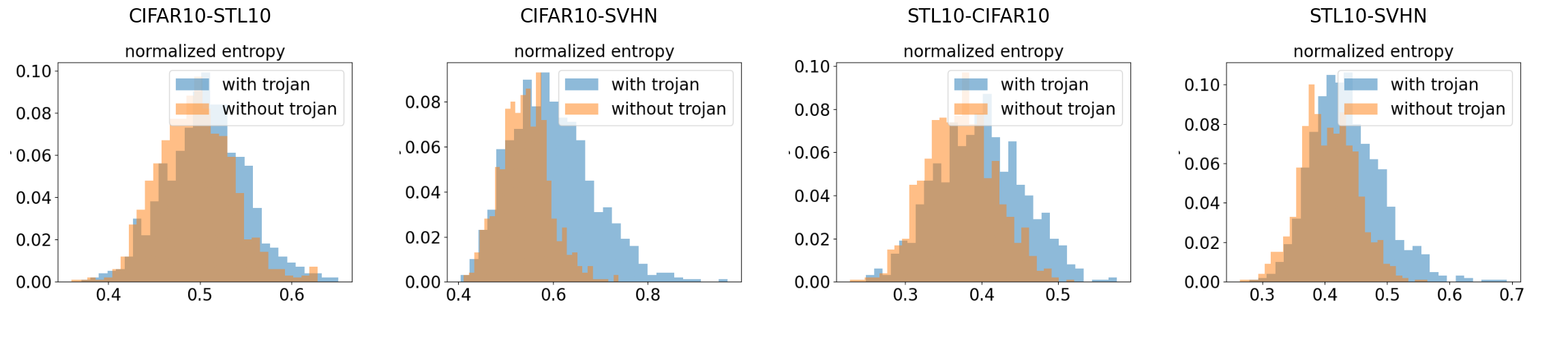}
    \caption{Resilient to STRIP~\cite{gao2019strip}. INACTIVE can withstand the STRIP runtime defense.}
    \label{strip}
\end{figure*}

\noindent
\textbf{STRIP.}\
STRIP~\cite{gao2019strip} scrutinizes a suspect sample by overlaying diverse image patterns onto it and monitoring the uniformity of the model's predictions for these altered inputs. A low entropy score, suggesting consistent predictions across these perturbed samples, would lead STRIP to classify the sample as Trojan-infected. The experimental outcomes depicted in \autoref{strip} reveal that the entropy values for clean and Trojan-compromised models produced through our method overlap considerably, signifying that our attack can withstand the STRIP runtime defense. INACTIVE manages to evade STRIP as the superimposition process destroys the attack's trigger, thereby causing both the Trojan-induced and clean samples to exhibit significant alterations in prediction when superimposed, aligning them with what is observed in clean cases.

\noindent \textbf{Adaptive Defense.} \
Our threat model assumes the attacker controls pre-downstream training, while the defender can only intervene during downstream training and inference. 
Since our trigger injection operates in the HSV color space, we introduce Salt and Pepper Noise and Poisson Noise perturbations in the HSV color space of both downstream training and inference phase, simulating potential adaptive defense strategies a defender might use to disrupt the attack.
We use CIFAR10 as the pre-training dataset and GTSRB, STL10, and SVHN as downstream datasets.
Experimental results show INACTIVE’s ASR remains unaffected, demonstrating its robustness. 
This resilience stems from our disentanglement loss, which increases the distributional gap between backdoor and augmented samples. This design inherently enhances the robustness of the trigger, as the larger distributional gap in the color space ensures that the trigger remains distinguishable even when noise is introduced to the samples. Consequently, the trigger is less affected by noisy samples and retains its effectiveness in activating the backdoor.

\begin{table}[htbp]
    \centering
    \scriptsize
\begin{tabular}{|c|c|c|c|c|}
\hline
\multirow{2}{*}{Downstream dataset} & \multicolumn{2}{c|}{\makecell[c]{Salt and Pepper on HSV}} & \multicolumn{2}{c|}{\makecell[c]{Poisson Noise on HSV}} \\
\cline{2-5}
& BA & ASR & BA & ASR \\
\hline
GTSRB & 81.96\% & 100\% & 81.81\% & 100\% \\
\hline
STL10 & 73.81\% & 99.38\% & 73.65\% & 99.41\% \\
\hline
SVHN & 60.17\% & 95.34\% & 60.65\% & 91.11\% \\
\hline
\end{tabular}
\caption{Adaptive defense evaluation results pre-trained on CIFAR10.}
\label{tab:Adaptive defense}
\end{table}
\subsection{Sensitivity Evaluation}
\label{supple:sensitive}
\begin{figure*}[t]
    \centering
    \includegraphics[width=0.7\textwidth]{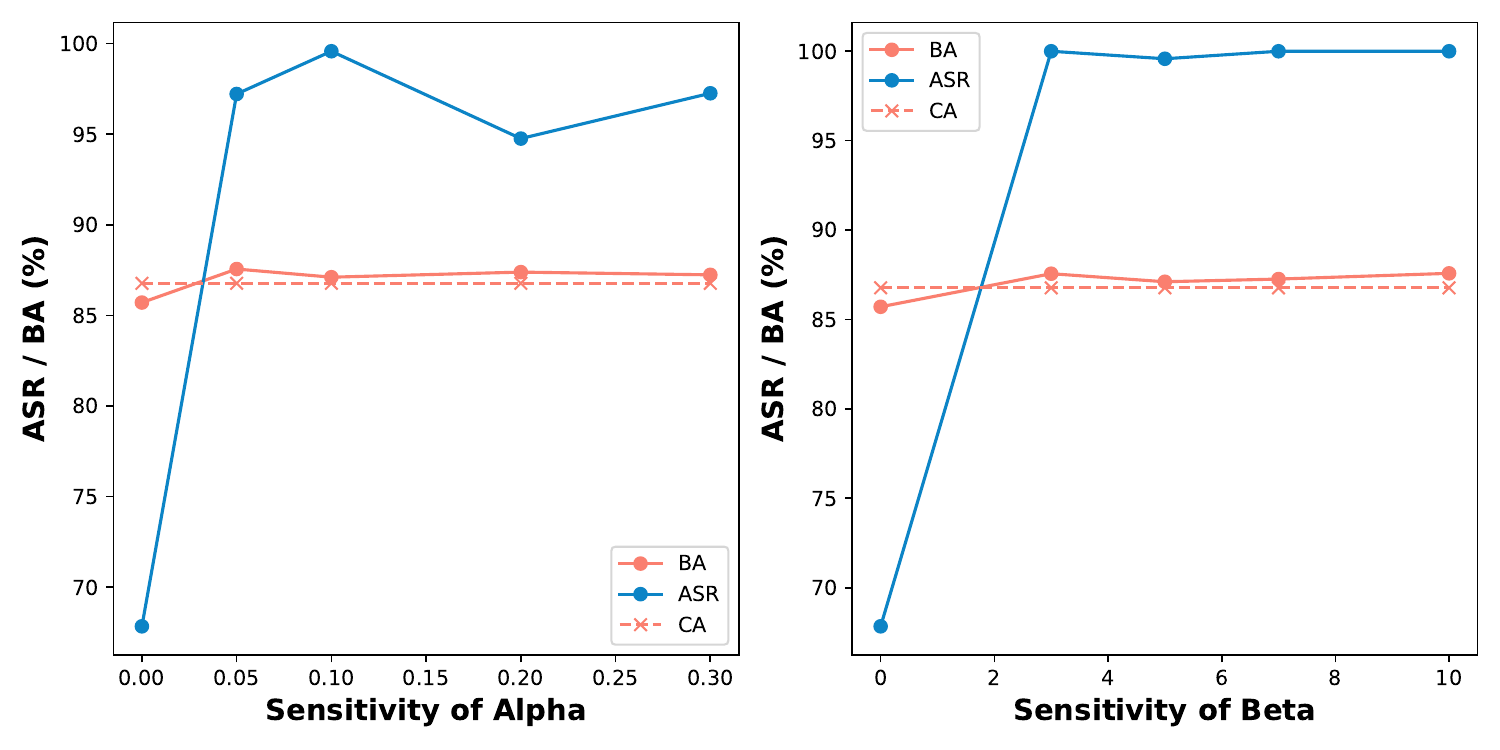}
    \caption{Sensitivity evaluation of hyper-parameters. INACTIVE performs consistently over these parameter ranges.}
    \label{fig:sensitive}
\end{figure*}

\autoref{fig:sensitive} illustrates the sensitivity evaluation of the hyper-parameters alpha and beta, showcasing that the model's performance exhibits relative insensitivity to parameter variations. Specifically, for alpha, both the Attack Success Rate (ASR) and Backdoored Accuracy (BA) remain relatively constant across a range of alpha values from 0.05 to 0.30, suggesting that once an optimal threshold is reached, the model's performance is stable despite further adjustments to alpha. Similarly, for beta, changes across a broader spectrum, from 0 to 10, show that ASR quickly stabilizes at 100\%, while BA and CA display minimal fluctuations, indicating the robust model performance that is not significantly affected by variations in beta. The horizontal axis in both graphs represents the incremental values of alpha and beta, demonstrating the model's consistent performance over these parameter ranges.

\subsection{Results Analysis}
\label{supple:Results Analysis}

\begin{figure*}[t]
    \centering
    \includegraphics[width= .8\textwidth]{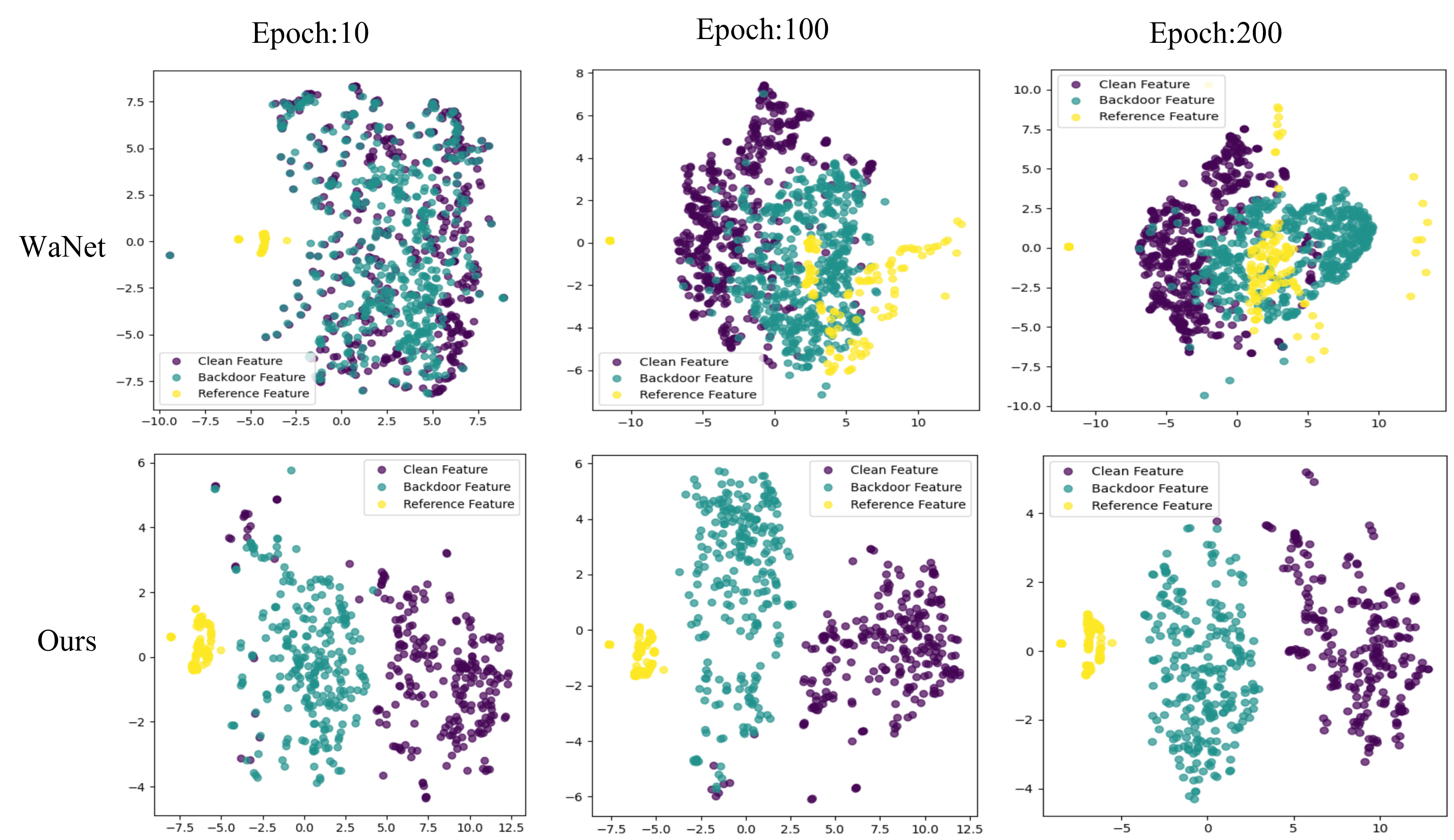}
    \caption{Visualization of feature space segregation over epochs demonstrates the efficacy of our backdoor trigger in creating a distinguishable separation between backdoor and clean features, unlike WaNet~\cite{nguyen2021wanet} where the features remain intermingled, highlighting our method's utility in securing encoders against backdoor threats.}
    \label{epoch}
\end{figure*}

\autoref{epoch} presents a visual analysis of the feature shifts induced by backdoor injection across different epochs for our proposed method compared to the WaNet~\cite{nguyen2021wanet}. Throughout the backdoor injection process, it becomes evident that our method's trigger effectively separates the backdoor features from the clean features. This distinction is observable as the training progresses from epoch 10 through to epoch 200, where the separation between the backdoor and clean features becomes increasingly pronounced, indicating a clear demarcation that our encoder can exploit to differentiate between the two.

In contrast, the bottom row representing WaNet's performance shows a significant overlap of backdoor and clean features. Even at epoch 200, the features remain intermixed, demonstrating WaNet's inability to distinguish effectively between the two, as the clean, backdoor, and reference features are all entangled.
The effectiveness of our backdoor trigger is thus underscored. It introduces a discernible feature shift that an encoder can leverage to recognize backdoored inputs, validating the practical utility of our approach in enhancing model security against backdoor attacks.

To illustrate distribution shifts during SSL backdoor injection, we use STL10 for pre-training and compute KL divergence between the distributions of clean, backdoor, and reference samples. Higher $\text{KL}(\text{Clean} \,||\, \text{Backdoor})$ indicates greater separation, while lower $\text{KL}(\text{Backdoor} \,||\, \text{Reference})$ shows alignment with the target. 
Both metrics validate our trigger's effectiveness. 
Results below demonstrate with epochs increasing, our backdoor trigger gradually creates a distinguishable separation between clean and backdoor features, and a closer distance between target and backdoor features, aligning with the conclusions in \autoref{epoch} feature space segregation visualization over epochs.

\begin{table}[ht]
\centering
\scriptsize
\begin{tabular}{|c|c|c|c|c|}
\hline
Epoch & 10      & 20      & 50      & 100     \\ \hline
$\text{KL}(\text{Clean} \,||\, \text{Backdoor})$  & 12.803  & 17.037 & 18.671 & 19.028 \\ \hline
$\text{KL}(\text{Backdoor} || \text{Reference})$ & 0.5034  & 0.0591  & 0.0071  & 0.0017  \\ \hline
\end{tabular}
\caption{KL divergence values across different epochs.}
\label{tab:kl_divergence}
\end{table}

\subsection{Stealthiness Evaluation}
\subsubsection{Algorithmic Metrics}

\autoref{tab:stealthiness_detail} and \autoref{tab:stealthiness_imagenet} show the detailed experimental results of the stealthy metrics across various backdoor attack methods and datasets.

\begin{table*}[t]
\scriptsize
\centering
\begin{tabular}{c c cccccc}
\toprule
Model & \makecell{Pre-training Dataset} & \makecell{Downstream Dataset} & SSIM$\uparrow$ & PSNR$\uparrow$ & LPIPS$\downarrow$ & FSIM$\uparrow$ & FID$\downarrow$ \\
\midrule
\multirow{7}{*}{\makecell{Badencoder~\cite{Jia_Liu_Gong_2022}\\/DRUPE~\cite{tao2023distribution}}} & \multirow{3}{*}{CIFAR10} & STL10 & 0.8375 & 14.3745 & 0.03894 & 0.7788 & 63.7301 \\
 &  & SVHN & 0.8380 & 15.0739 & 0.11092 & 0.8518 & 33.1811 \\
 &  & GTSRB & 0.8309 & 12.8845 & 0.08094 & 0.8295 & 63.1784 \\ \cmidrule{2-8}
 & \multirow{3}{*}{STL10} & CIFAR10 & 0.8455 & 14.9113 & 0.03696 & 0.8598 & 13.8215 \\
 &  & SVHN & 0.8407 & 15.0849 & 0.11110 & 0.8165 & 64.5712 \\
 &  & GTSRB & 0.8308 & 12.8965 & 0.08597 & 0.8306 & 63.2014 \\ 
 \cmidrule{2-8}
 & \textbf{Average} &  & \textbf{0.8372} & \textbf{14.2043} & \textbf{0.07747} & \textbf{0.8278} & \textbf{50.2806} \\
\midrule 
\multirow{7}{*}{CTRL~\cite{Li_2023_ICCV}} & \multirow{3}{*}{CIFAR10} & STL10 & 0.9415 & 32.4195 & 0.00017 & 0.9115 & 47.7871 \\
 &  & SVHN & 0.8695 & 32.3533 & 0.00065 & 0.7032 & 157.6731  \\
 &  & GTSRB & 0.8965 & 32.4565 & 0.00021 & 0.9787 & 7.9540  \\ 
 \cmidrule{2-8}
 & \multirow{3}{*}{STL10} & CIFAR10 & 0.9356 & 32.4603 & 0.00019 & 0.9827 & 11.9283  \\
 &  & SVHN & 0.8721 & 32.4198 & 0.00062 & 0.9347 & 59.4134  \\
 &  & GTSRB & 0.8945 & 32.5226 & 0.00024 & 0.9773 & 8.8889  \\ 
 \cmidrule{2-8}
 & \textbf{Average} &  & \textbf{0.9016} & \textbf{32.4387} & \textbf{0.00035} & \textbf{0.9147} & \textbf{48.9408} \\
\midrule
\multirow{7}{*}{WaNet~\cite{nguyen2021wanet}} & \multirow{3}{*}{CIFAR10} & STL10 & 0.7378 & 14.3472 & 0.04657 & 0.6611 & 97.1471 \\
 &  & SVHN & 0.7941 & 15.2503 & 0.10023 & 0.5927 & 129.6770  \\
 &  & GTSRB & 0.7792 & 13.1140 & 0.07615 & 0.7320 & 67.4510  \\ 
 \cmidrule{2-8}
 & \multirow{3}{*}{STL10} & CIFAR10 & 0.7728 & 14.9430 & 0.03996 & 0.6510 & 96.0399  \\
 &  & SVHN & 0.8026 & 15.4044 & 0.09544 & 0.5924 & 129.8306  \\
 &  & GTSRB & 0.7687 & 13.1645 & 0.08358 & 0.7314 & 67.7673  \\ 
 \cmidrule{2-8}
 & \textbf{Average} &  & \textbf{0.7759} & \textbf{14.3706} & \textbf{0.07366} & \textbf{0.6601} & \textbf{97.9855} \\
\midrule
\multirow{7}{*}{Ins-Kelvin~\cite{liu2019abs}} & \multirow{3}{*}{CIFAR10} & STL10 & 0.4427 & 15.7024 & 0.11110 & 0.7011 & 107.4607  \\
 &  & SVHN & 0.4213 & 15.4081 & 0.20070 & 0.5905 & 121.0396  \\
 &  & GTSRB & 0.6224 & 17.4669 & 0.10820 & 0.7393 & 60.8467  \\ 
 \cmidrule{2-8}
 & \multirow{3}{*}{STL10} & CIFAR10 & 0.4624 & 15.6372 & 0.10970 & 0.6582 & 85.2866  \\
 &  & SVHN & 0.4358 & 15.4436 & 0.19840 & 0.6342 & 130.1333  \\
 &  & GTSRB & 0.5942 & 17.5105 & 0.11640 & 0.7422 & 62.0004  \\ 
 \cmidrule{2-8}
 & \textbf{Average} &  & \textbf{0.4965} & \textbf{16.1948} & \textbf{0.14075} & \textbf{0.6776} & \textbf{94.4612} \\
\midrule
\multirow{7}{*}{Ins-Xpro2~\cite{liu2019abs}} & \multirow{3}{*}{CIFAR10} & STL10 & 0.5672 & 17.8706 & 0.03451 & 0.7806 & 66.9640 \\
 &  & SVHN & 0.5324 & 17.4862 & 0.06699 & 0.7698 & 27.4974  \\
 &  & GTSRB & 0.6947 & 18.3951 & 0.03151 & 0.8997 & 10.7916 \\ 
 \cmidrule{2-8}
 & \multirow{3}{*}{STL10} & CIFAR10 & 0.5354 & 17.9291 & 0.03806 & 0.8063 & 19.0678  \\
 &  & SVHN & 0.5057 & 17.4956 & 0.07152 & 0.7505 & 45.3211  \\
 &  & GTSRB & 0.6541 & 18.4070 & 0.03632 & 0.8932 & 11.9285  \\ 
 \cmidrule{2-8}
 & \textbf{Average} &  & \textbf{0.5816} & \textbf{17.7640} & \textbf{0.04615} & \textbf{0.8167} & \textbf{30.2617} \\
\midrule
POIENC~\cite{liu2022poisonedencoder} & CIFAR10 & CIFAR10 & 0.1214 & 11.2787 & 0.15867 & 0.5967 & 172.2200  \\
\midrule
BLTO~\cite{sun2024backdoor} & CIFAR10 & CIFAR10 & 0.8417 & 29.6756 & 0.00941 & 0.9501 & 36.3848  \\
\midrule
SSLBKD~\cite{Saha_Tejankar_Koohpayegani_Pirsiavash_2022} & CIFAR10 & CIFAR10 & 0.8737 & 16.2414 & 0.09640 & 0.8913 & 118.3200  \\
\midrule
\multirow{7}{*}{Ours} & \multirow{3}{*}{CIFAR10} & STL10 & 0.9405 & 24.0804 & 0.02311 & 0.9161 & 45.8759 \\
 &  & SVHN & 0.9807 & 46.0749 & 0.00040 & 0.9976 & 1.8810 \\
 &  & GTSRB & 0.9687 & 37.4394 & 0.00338 & 0.9944 & 1.2025  \\ 
 \cmidrule{2-8}
 & \multirow{3}{*}{STL10} & CIFAR10 & 0.9928 & 46.5797 & 0.00019 & 0.9981 & 0.5018  \\
 &  & SVHN & 0.9847 & 45.7410 & 0.00031 & 0.9464 & 30.0092  \\
 &  & GTSRB & 0.9905 & 46.5149 & 0.00011 & 0.9977 & 0.2157  \\ 
 \cmidrule{2-8}
 & \textbf{Average} &  & \textbf{0.9763} & \textbf{41.0717} & \textbf{0.0046} & \textbf{0.9751} & \textbf{13.2810} \\
\bottomrule
\end{tabular}%
\caption{Stealthiness Metrics across Methods, Pre-training and Downstream Datasets.}
\label{tab:stealthiness_detail}
\end{table*}

\begin{table}[t]
\scriptsize
\centering
\resizebox{\linewidth}{!}{%
\begin{tabular}{c c cccc}
\toprule
Model & \makecell{Pre-training\\ Dataset} & \makecell{Downstream\\ Dataset} & SSIM & PSNR(dB) & LPIPS \\
\midrule
\multirow{4}{*}{ISSBA~\cite{li2021invisible}} & \multirow{3}{*}{ImageNet} & STL10 & 0.7836 & 30.3333 & 0.05615 \\
 &  & GTSRB & 0.7292 & 31.7344 & 0.11651 \\
 &  & SVHN & 0.6860 & 31.9810 & 0.20006 \\ \cmidrule{2-6}
 & \textbf{Average} &  & \textbf{0.7329} & \textbf{31.3496} & \textbf{0.12424} \\
\midrule
\multirow{4}{*}{Ours} 
 & \multirow{3}{*}{ImageNet} & STL10 & 0.9900 & 38.3300 & 0.00400 \\
 &  & GTSRB & 0.9900 & 32.1400 & 0.01700 \\
 &  & SVHN & 0.9800 & 33.2500 & 0.01600 \\ \cmidrule{2-6}
 & \textbf{Average} &  & \textbf{0.9867} & \textbf{34.5733} & \textbf{0.01233} \\
\bottomrule
\end{tabular}%
}
\caption{Stealthiness Metrics across Methods on ImageNet.}
\label{tab:stealthiness_imagenet}
\end{table}

\subsubsection{Human Inspection}
\begin{figure*}
    \centering
    \includegraphics[width=1\linewidth]{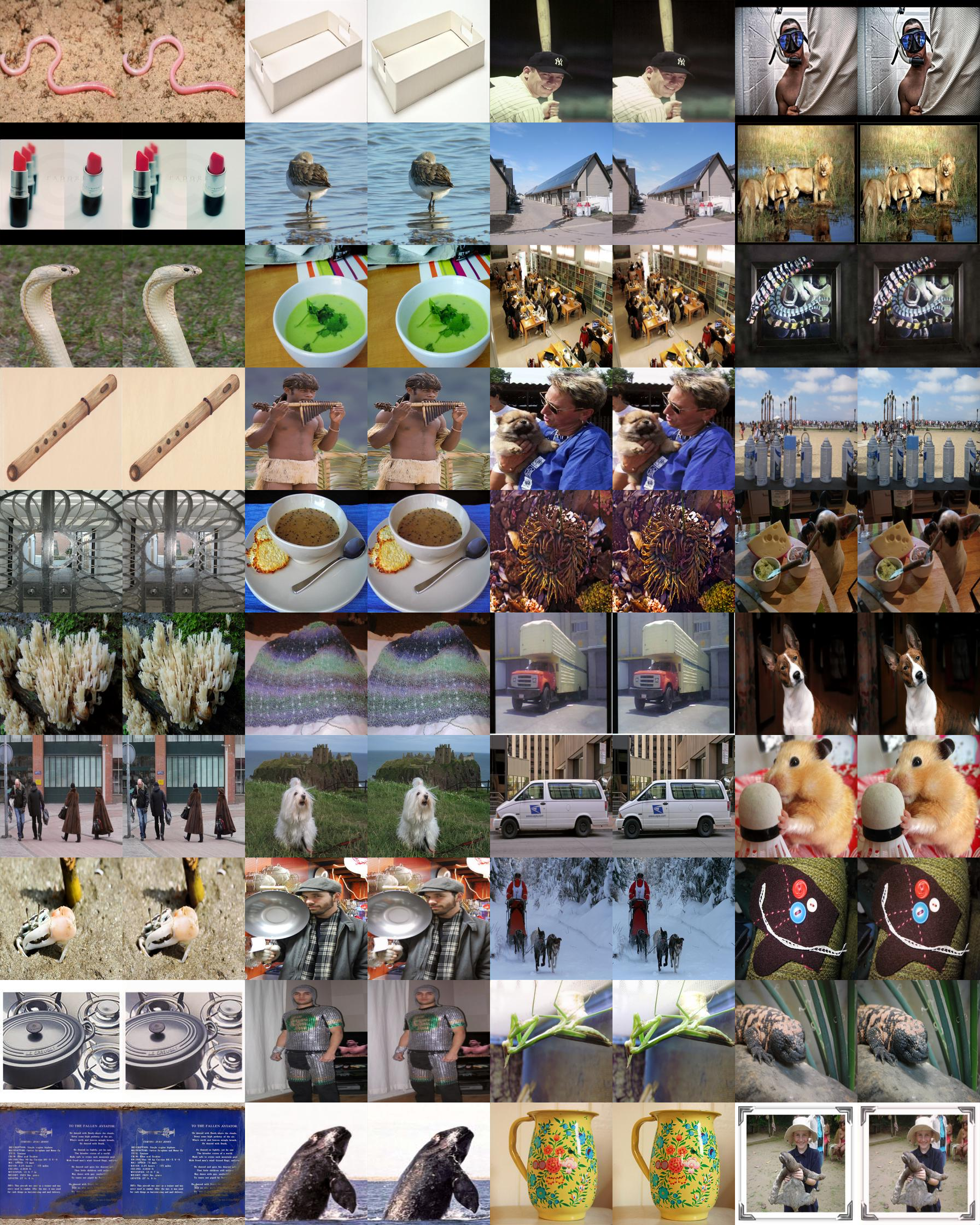}
    \caption{More example ImageNet images with triggers generated by INACTIVE. The backdoor image is on the left side of the same set of images and the clean image is on the right.}
    \label{fig:combined_image}
\end{figure*}

We conducted a human inspection study~\cite{nguyen2021wanet} to evaluate the stealthiness of various backdoor attacks. 
We introduce the detailed settings in \autoref{supple:train_detail}.
The results in \autoref{tab:human_inspection} indicate that traditional methods like SSLBKD, BadEncoder/DRUPE, and POIENC have relatively low fooling rates, with SSLBKD at just 4.9\% overall, suggesting they are easier for humans to detect.
In contrast, our method achieves a higher fooling rate of 49.5\% overall, indicating much greater stealth. However, as shown in \autoref{fig:combined_image}, subtle artifacts, such as slight color changes, can still be detected by humans.

\begin{table}[]
\scriptsize
\setlength\tabcolsep{1pt}
\centering
\begin{tabular}{ccccccccc}
\toprule
Images & SSLBKD & CTRL & \makecell{BadEncoder\\/DRUPE} & POIENC & Ins-Kelvin & BLTO & ISSBA & \textbf{Ours} \\
\midrule
Backdoor & 2.4 & 31.2 & 4.0 & 9.6 & 12.8 & 22.4 & 24.8 & \textbf{50.6} \\
Clean & 7.4 & 27.2 & 8.8 & 16.8 & 10.4 & 27.6 & 26.4 & \textbf{48.4} \\
Both & 4.9 & 29.2 & 6.4 & 13.2 & 11.6 & 25.0 & 25.6 & \textbf{49.5} \\
\bottomrule
\end{tabular}

\caption{Success fooling rates$\uparrow$ (\%) of different methods. Our method achieves the greatest stealth with the highest overall success fooling rates.}
\label{tab:human_inspection}
\vspace{-5pt}
\end{table}

\section{Implementation Details}
\label{supple:train_detail}
The following sections provide a comprehensive breakdown of
the process of implementing the attack, which is methodically organized into three distinct phases: pre-training the encoder, injecting the backdoor, and training the downstream classifiers. Additionally, we explain the HSV\&HSL Color Spaces and Wasserstein Distance used in the method in detail.  Moreover, we provide more details on the datasets and experimental setup details.

\subsection{Pre-training Image Encoders}
\label{supple:pretrain_encoder}
The initial step involves pre-training an image encoder using a specific dataset, hereinafter referred to as the pre-training dataset. In alignment with the experimental framework established in BadEncoder~\cite{Jia_Liu_Gong_2022}, we opt for CIFAR10 or STL10 as our pre-training datasets. These choices are informed by their relatively large size and superior data quality. Crucially, our experimentation leverages the pre-trained weights from BadEncoder's image encoders, which underwent training over 1,000 epochs employing the Adam optimizer at a learning rate of 0.001~\cite{Jia_Liu_Gong_2022}. Specifically, for the CIFAR10 dataset, the encoder is pre-trained using only the training images, excluding their labels. Meanwhile, for STL10, the pre-training also incorporates its additional unlabeled images.

\begin{table*}[t]
\centering
\scriptsize
\begin{tabular}{@{}cccccccccc@{}}
\toprule
\textbf{Module} & \textbf{Layer} & \textbf{Type} & \textbf{Kernel Size} & \textbf{Stride} & \textbf{Padding} & \textbf{Channel I/O} & \textbf{Normalization} & \textbf{Activation} & \textbf{Input} \\ \midrule
\multirow{6}{*}{\textbf{Encoder}} & Maxpool & MaxPool2d & 2x2 & 2 & 0 & - & - & - & Input Image \\
 & Conv1 & Conv2d & 3x3 & 1 & 1 & 3/8 & BatchNorm2d (8) & ReLU & Maxpool \\
 & Conv2 & Conv2d & 3x3 & 1 & 1 & 8/16 & BatchNorm2d (16) & ReLU & Conv1 \\
 & Conv3 & Conv2d & 3x3 & 1 & 1 & 16/32 & BatchNorm2d (32) & ReLU & Conv2 \\
 & Conv4 & Conv2d & 3x3 & 1 & 1 & 32/64 & BatchNorm2d (64) & ReLU & Conv3 \\
 & Conv5 & Conv2d & 3x3 & 1 & 1 & 64/128 & BatchNorm2d (128) & ReLU & Conv4 \\ \midrule
\multirow{9}{*}{\textbf{Decoder}} & Up5 & Upsample + Conv2d & -/3x3 & -/1 & -/1 & 128/64 & BatchNorm2d (64) & ReLU & Conv5 \\
 & Up\_conv5 & Conv2d & 3x3 & 1 & 1 & 128/64 & BatchNorm2d (64) & ReLU & Up5 \\
 & Up4 & Upsample + Conv2d & -/3x3 & -/1 & -/1 & 64/32 & BatchNorm2d (32) & ReLU & Up\_conv5 \\
 & Up\_conv4 & Conv2d & 3x3 & 1 & 1 & 64/32 & BatchNorm2d (32) & ReLU & Up4 \\
 & Up3 & Upsample + Conv2d & -/3x3 & -/1 & -/1 & 32/16 & BatchNorm2d (16) & ReLU & Up\_conv4 \\
 & Up\_conv3 & Conv2d & 3x3 & 1 & 1 & 32/16 & BatchNorm2d (16) & ReLU & Up3 \\
 & Up2 & Upsample + Conv2d & -/3x3 & -/1 & -/1 & 16/8 & BatchNorm2d (8) & ReLU & Up\_conv3 \\
 & Up\_conv2 & Conv2d & 3x3 & 1 & 1 & 16/8 & BatchNorm2d (8) & ReLU & Up2 \\
 & Conv\_1x1 & Conv2d & 1x1 & 1 & - & 8/3 & - & - & Up\_conv2 \\ \bottomrule
\end{tabular}%
\caption{The detailed network structure of backdoor injector.}
\label{unet}
\end{table*}

\subsection{Backdoor Injection}
\label{supple: Backdoor Injection}
The subsequent phase introduces a backdoor into the pre-trained encoder. This is achieved through the use of Shadow Dataset, which constitutes a smaller subset of the pre-training dataset. Consistent with preceding studies~\cite{Jia_Liu_Gong_2022}, this shadow dataset comprises 50,000 randomly selected training samples. Additionally, we employ the same reference samples as those used in BadEncoder~\cite{Jia_Liu_Gong_2022}. The backdoor is injected using a designated algorithm, which leverages a combination of $\mathcal{L}_{\text {consistency}}$, $\mathcal{L}_{\text {utility}}$, and $\mathcal{L}_{\text {alignment}}$. The notations are the same as the main paper.

\noindent
\textbf{Consistency Loss.}\
The consistency loss decreases when the feature vectors generated by the backdoored image encoder and the clean image encoder for the reference inputs are more alike. 
\begin{equation}
\label{L_consistency}
\mathcal{L}_{\text {consistency}}=-\frac{\sum_{i=1}^t \sum_{j=1}^{r_i} s\left(\mathcal{F}_\theta^{\prime}\left(\boldsymbol{x}_{i j}\right), \mathcal{F}_\theta\left(\boldsymbol{x}_{i j}\right)\right)}{\sum_{i=1}^t r_i}
\end{equation}

\noindent
\textbf{Utility Loss.}\
Utility loss diminishes when the backdoored image encoder and the clean image encoder yield more closely matching feature vectors for a clean input within the shadow dataset.
\begin{equation}
\label{L_utility}
\mathcal{L}_{\text {utility}}=-\frac{1}{\left|\mathcal{D}_s\right|} \cdot \sum_{\boldsymbol{x} \in \mathcal{D}_s} s\left(\mathcal{F}_\theta^{\prime}(\boldsymbol{x}), \mathcal{F}_\theta(\boldsymbol{x})\right)
\end{equation}
\subsection{Training Downstream Classifiers}
\label{supple:Training Downstream Classifiers}
Finally, using the pre-trained image encoder, we proceed to train classifiers for three separate datasets, hereafter referred to as downstream datasets. Our approach utilizes a two-layer fully connected neural network as the classifier architecture, with the first and second layers consisting of 512 and 256 neurons, respectively. The testing subset of each downstream dataset is utilized for evaluation purposes. The training of the classifier is conducted over 500 epochs using the cross-entropy loss function and the Adam optimizer, with a set learning rate of 0.0001.

\subsection{Technical Background}
\label{supple:Technical Background}
\noindent
\textbf{HSV Color Space.}\
HSV and HSL color spaces align more closely with human color perception, unlike RGB's additive mixing. They allow for intuitive adjustments of tint, shade, and tone through single-dimension modifications. 
They have wide applications in the computer vision domain~\cite{Liu_Shu_Pan_Shi_Han_2023,Dou_Wang_Li_Wang_Li_Liu_2021}, as detailed in \autoref{supple:Technical Background}.

The HSV color model describes color based on three primary attributes: hue, saturation, and value. Hue (H) represents the color itself, essentially the aspect of color we typically refer to by names like red or yellow. Saturation (S) indicates the vividness or richness of the color. A higher saturation means a more intense, pure color. Value (V), often referred to as brightness, measures the lightness or darkness of the color~\cite{Shuhua_Gaizhi_2010}. HSV color space can be converted from RGB color space through \autoref{convert1}, \autoref{convert2}, \autoref{convert3}, \autoref{convert4}~\cite{Saravanan_Yamuna_Nandhini_2016}.

\noindent
\textbf{HSL Color Space.}\
The HSL color model is defined by three cylindrical coordinates: hue (H), saturation (S), and lightness (L), which together capture the subtleties of color. HSL color space can be converted from RGB color space through \autoref{convert1}, \autoref{convert2}, \autoref{convert5}, \autoref{convert6}~\cite{Saravanan_Yamuna_Nandhini_2016}.

\begin{equation}
\label{convert1}
\begin{aligned}
& \mathrm{R}^{\prime} = \mathrm{R} / 255; \\
& \mathrm{G}^{\prime} = \mathrm{G} / 255; \\
& \mathrm{B}^{\prime} = \mathrm{B} / 255; \\
& \mathrm{C}_{\max} = \operatorname{MAX}\left(\mathrm{R}^{\prime}, \mathrm{G}^{\prime}, \mathrm{B}^{\prime}\right); \\
& \mathrm{C}_{\min} = \operatorname{MIN}\left(\mathrm{R}^{\prime}, \mathrm{G}^{\prime}, \mathrm{B}^{\prime}\right); \\
& \Delta = \mathrm{C}_{\max} - \mathrm{C}_{\min}; \\
\end{aligned}
\end{equation}

\begin{equation}
\label{convert2}
\begin{aligned}
\mathrm{H} = \left\{
\begin{array}{l}
60^{\circ} \times\left(\frac{\mathrm{G}^{\prime} - \mathrm{B}^{\prime}}{\Delta} \bmod 6\right), \quad \mathrm{C}_{\max} = \mathrm{R}^{\prime} \\
60^{\circ} \times\left(\frac{\mathrm{B}^{\prime} - \mathrm{R}^{\prime}}{\Delta} + 2\right), \quad \mathrm{C}_{\max} = \mathrm{G}^{\prime} \\
60^{\circ} \times\left(\frac{\mathrm{R}^{\prime} - \mathrm{G}^{\prime}}{\Delta} + 4\right), \quad \mathrm{C}_{\max} = \mathrm{B}^{\prime}
\end{array}
\right.
\end{aligned}
\end{equation}

\begin{equation}
    \mathrm{S}=\left\{\begin{array}{cc}
0 \quad, & \Delta=0 \\
\frac{\Delta}{\mathrm{C}_{\max }}, & \Delta \neq 0
\end{array}\right.
\label{convert3}
\end{equation}

\begin{equation}
    \mathrm{V} = \mathrm{C}_{\max }
\label{convert4}
\end{equation}

\begin{equation}
    \mathrm{S}=\left\{\begin{array}{cl}
0, & \Delta=0 \\
\frac{\Delta}{1-|2\mathrm{L}-1|} & , \Delta \neq 0
\end{array}\right.
\label{convert5}
\end{equation}

\begin{equation}
    \mathrm{L}=\left(\mathrm{C}_{\max }+\mathrm{C}_{\min }\right) / 2
\label{convert6}
\end{equation}






\noindent
\textbf{Wasserstein Distance.}\
Wasserstein distance is commonly employed for quantifying differences between two latent distributions without known common support or density functions~\cite{tao2023distribution}. Specifically, we focus on the 2-Wasserstein distance, which is defined below.
\begin{equation}
    \mathcal{W}(\zeta, \tau)=\left(\inf _{\psi \in \Pi(\zeta, \tau)} \int_{(u, v) \sim \psi} p(u, v)\|u-v\|_2 d u d v\right)^{1 / 2},
\end{equation}
where \(\xi\) and \(\tau\) correspond to the marginal probability distributions of the clean and poisoned datasets. \(\psi\) denotes the combined distribution of both clean and poisoned data samples. \(\inf\) represents the infimum, which is the lowest boundary value for the calculated distances within the scope of the joint distribution \(\psi\). The integral part of the equation aggregates the distances between each pair of data points, \( (u, v) \), where one belongs to the clean set and the other to the poisoned set, as drawn from the joint distribution \(\psi\). Here, \( p(u, v) \) defines the probability of concurrently selecting both data samples. The term \( ||u - v||_2 \) measures the \( L^2 \) distance between the two data points~\cite{villani2009optimal}.

\subsection{Experiments Details}
\label{supple:exp_detail}
\begin{figure}[ht]
    \centering
    \begin{minipage}[b]{0.11\textwidth}
        \centering
        \includegraphics[width=\textwidth]{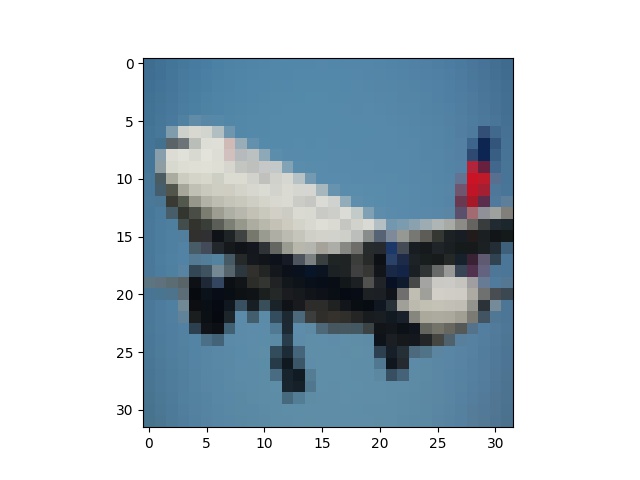}
        \caption*{(a) CIFAR10}
    \end{minipage}
    \centering
    \begin{minipage}[b]{0.11\textwidth}
        \centering
        \includegraphics[width=\textwidth]{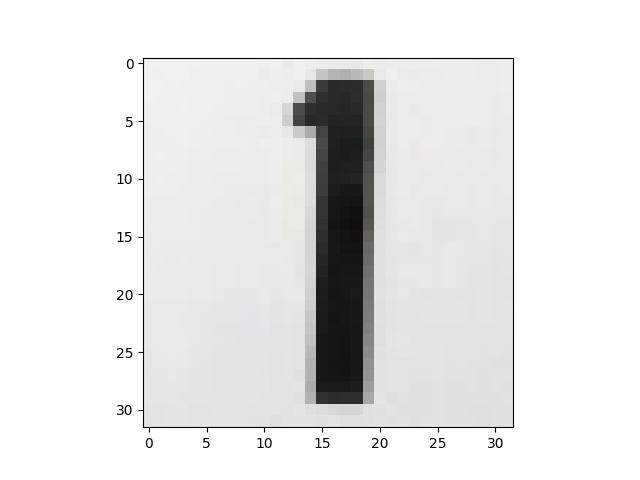}
        \caption*{(b) SVHN}
    \end{minipage}
    \hfill
    \begin{minipage}[b]{0.11\textwidth}
        \centering
        \includegraphics[width=\textwidth]{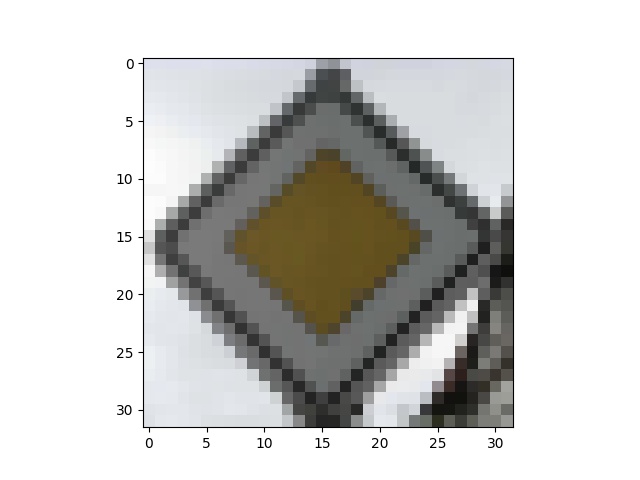}
        \caption*{(c) GTSRB}
    \end{minipage}
    \hfill
    \begin{minipage}[b]{0.11\textwidth}
        \centering
        \includegraphics[width=\textwidth]{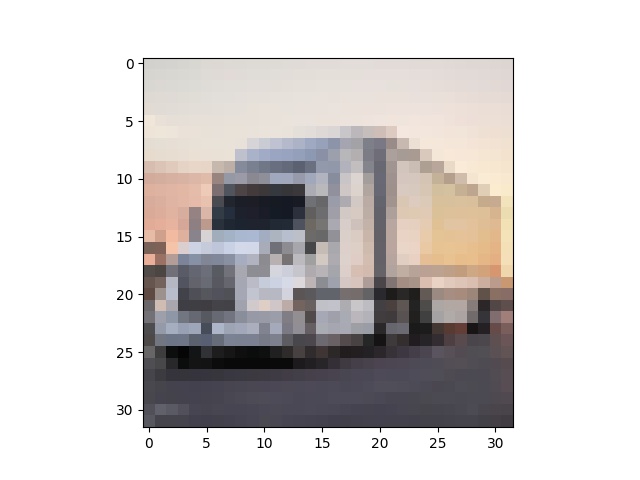}
        \caption*{(d) STL10}
    \end{minipage}
    \caption{The default reference inputs for CIFAR10 SVHN, GTSRB, and STL10.}
    \label{fig:three_images}
\end{figure}

\begin{figure}[ht]
    \centering
    \begin{minipage}[b]{0.15\textwidth}
        \centering
        \includegraphics[width=\textwidth]{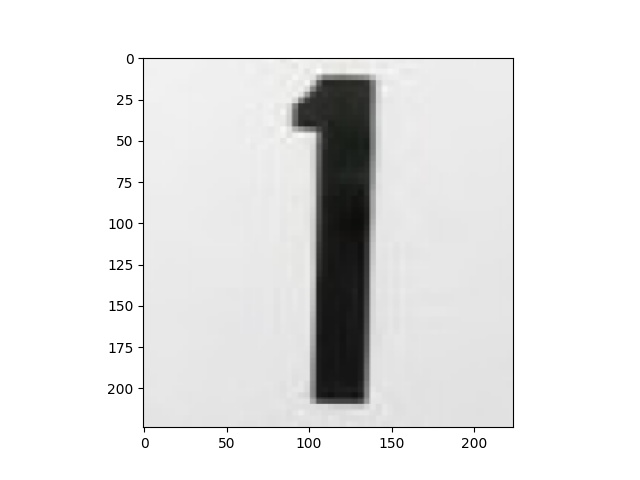}
        \caption*{(a) SVHN}
    \end{minipage}
    \hfill
    \begin{minipage}[b]{0.15\textwidth}
        \centering
        \includegraphics[width=\textwidth]{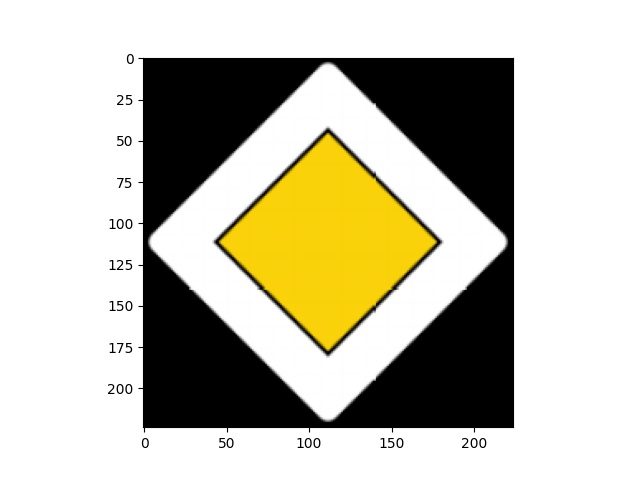}
        \caption*{(b) GTSRB}
    \end{minipage}
    \hfill
    \begin{minipage}[b]{0.15\textwidth}
        \centering
        \includegraphics[width=\textwidth]{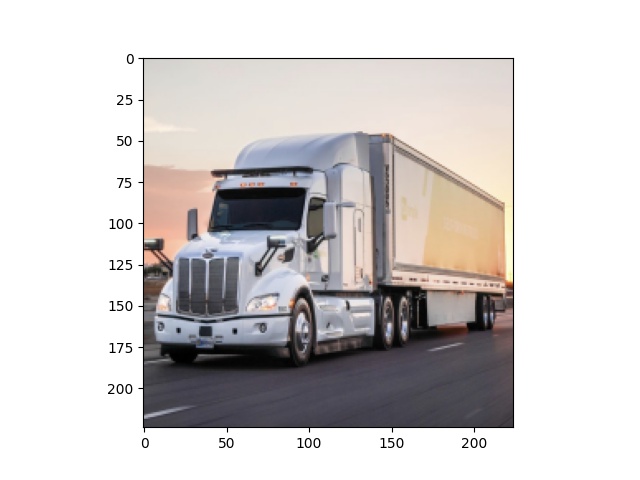}
        \caption*{(c) STL10}
    \end{minipage}
    \caption{The reference inputs for attacking the image encoder pre-trained on ImageNet by Google.}
    \label{resnet50_google_attack_input}
\end{figure}

\noindent
\textbf{Referece images.}\
Following BadEncoder~\cite{Jia_Liu_Gong_2022}, the reference images used in the experiments are shown in \autoref{fig:three_images} and \autoref{resnet50_google_attack_input}.

\noindent
\textbf{Dataset Details.}\
\label{supple:datasets}
We use the following datasets in our method evaluation.
\begin{itemize}
    \item  \textbf{CIFAR10}~\cite{krizhevsky2009learning}: This dataset comprises 60,000 images of 32×32×3 pixels and 10 different classes for basic image recognition tasks, divided into 50,000 for training and 10,000 for testing.
    \item \textbf{STL10}~\cite{pmlr-v15-coates11a}: STL10 includes 5,000 labeled training images and 8,000 for testing with a resolution of 96×96×3 pixels, across 10 classes. Additionally, it provides 100,000 unlabeled images for unsupervised learning. Notably, they are resized to 32×32×3 to be consistent with other datasets. 
    \item \textbf{GTSRB}~\cite{STALLKAMP2012323}: This dataset includes 51,800 images of traffic signs categorized into 43 classes. It is split into 39,200 training and 12,600 testing images, each sized at 32×32×3. 
    \item \textbf{SVHN}~\cite{37648}: SVHN is a dataset of digit images from house numbers in Google Street View, consisting of 73,257 training and 26,032 testing images, each 32×32×3 in size.
    \item \textbf{ImageNet}~\cite{russakovsky2015imagenet}: ImageNet is designed for large-scale object classification, featuring 1,281,167 training samples and 50,000 testing samples across 1000 categories. Each image has a resolution of 224x224 pixels with three color channels.
\end{itemize}

    

    

\noindent
\textbf{Parameter Setting Details.}\
\label{supple:Parameter Setting}
We consider a scenario where the attacker chooses a specific downstream task/dataset, a specific target class, and a specific reference input. For the datasets CIFAR10, STL10, GTSRB, and SVHN, the selected target classes are ``airplane'', ``truck'', ``priority sign'', and ``digit one'', respectively. 
CIFAR10 and STL10 are used as the default pre-training datasets. The shadow dataset following~\cite{Jia_Liu_Gong_2022} consists of 50000 images randomly chosen from the pre-training datasets. 
We adopt the default parameter settings as follows unless stated otherwise: 
\begin{enumerate}
    \item We set $\lambda_1 = 10, \lambda_2 = 0.025$ in the loss equation term \autoref{stealthy}, $\alpha=0.1, \beta=5$ in \autoref{algo1}, and $\mu=0.1$ in \autoref{algo2} to scale the loss terms to the same 0-1 range. 
    \item Cosine similarity is adopted to measure the similarity between two samples’ feature embeddings outputted by an encoder. 
    \item We adopt U-Net~\cite{ronneberger2015u} as the structure of the backdoor injector, which is widely used in image-to-image tasks.
    \item \autoref{algo1} and \autoref{algo2} utilize the Adam optimizer with 200 training epochs, a batch size of 256, and learning rates of 0.001 and 0.005 respectively. 
\end{enumerate}



\noindent
\textbf{Experimental Setup in Multi-modal Model.}\
Due to the unavailability of CLIP's original pre-training dataset, we utilize the training images from CIFAR-10, resized to 224×224×3, as our shadow dataset. We define $\alpha=0.1$ and $\beta=20$ in \autoref{algo1}, and $\mu=0.1$ in \autoref{algo2} to normalize the loss terms across similar scales. Additionally, \autoref{algo1} and \autoref{algo2} are configured with a batch size of 16, and learning rates are set at $10^{-6}$ and 0.005, respectively, for a duration of 100 epochs. SVHN serves as the downstream dataset with the target label ``one''. We employ the same reference inputs as described in~\cite{Jia_Liu_Gong_2022}. All other experimental settings remain consistent with those outlined in the Parameter Setting section of the main paper.

\noindent
\textbf{Experimental Setup in Real World Case ImageNet.}\
\label{supple:imagenet}
We selected a random 1\% subset of ImageNet's training images to serve as the shadow dataset. Additionally, we adjusted the size of each image in both the shadow and downstream datasets to dimensions of 224×224×3 as Google does in the encoder pre-training stage~\cite{chen2020simple}. 
We set $\alpha=0.1, \beta=20$ in \autoref{algo1}, and $\mu=0.1$ in \autoref{algo2} to scale the loss terms to the same range. Moreover, \autoref{algo1} and \autoref{algo2} here use a batch size of 16, and learning rates of $10^{-4}$ and 0.005 respectively.

\noindent
\textbf{Experimental Setup in Human Inspection Study.}
For each question, we randomly selected 50 images from the ImageNet dataset, generated backdoor variants for each attack method, and listed them in random order. Each method’s set included both original and backdoor images, totaling 100 images.
Before the survey, we thoroughly briefed each participant on backdoor triggers and baseline methods, confirming their understanding. We also manually reviewed each survey response for completeness.

To reduce bias, we included participants of diverse genders and ages. In all questions, images were presented in random order. Five trained participants then assessed each image, identifying it as either a backdoor or a clean sample.

To ensure consistency, we manually reviewed any outlier responses and asked participants additional questions on backdoor basics to confirm their task understanding.

\end{document}